%% file: main.tex
\documentclass[10pt,twocolumn,letterpaper]{article}

\usepackage[pagenumbers]{cvpr} 

\input{preamble}

\definecolor{cvprblue}{rgb}{0.21,0.49,0.74}
\usepackage[pagebackref,breaklinks,colorlinks,citecolor=cvprblue]{hyperref}

\usepackage[utf8]{inputenc} 
\usepackage[T1]{fontenc}    
\usepackage{url}            
\usepackage{booktabs}       
\usepackage{amsfonts}       
\usepackage{nicefrac}       
\usepackage{microtype}      
\usepackage{xcolor}         
\usepackage{subcaption}
\usepackage{multirow}
\usepackage{dblfloatfix}
\usepackage{wrapfig}
\usepackage{enumitem}
\usepackage{xr}
\usepackage[export]{adjustbox}
\usepackage[accsupp]{axessibility}
\usepackage{xcolor}
\usepackage[normalem]{ulem}
\useunder{\uline}{\ul}{}
\usepackage{axessibility}
\usepackage{graphicx}
\def\paperID{16214} 
\def\confName{CVPR}
\def\confYear{2024}

\title{On the Scalability of Diffusion-based Text-to-Image Generation}

\newcommand*{\affmark}[1][*]{\textsuperscript{#1}}

\author{ 
Hao Li\affmark[1,2], Yang Zou\affmark[1,2], Ying Wang\affmark[1,2],  Orchid Majumder\affmark[1,2], Yusheng Xie\affmark[1,2], R. Manmatha\affmark[1], \\Ashwin Swaminathan\affmark[1,2], Zhuowen Tu\affmark[1], Stefano Ermon\affmark[1], Stefano Soatto\affmark[1]\\
\affmark[1]AWS AI Labs, \affmark[2]Amazon AGI\\
{\tt\footnotesize \{haolimax, yanzo, lyiwang, orchid, yushx, manmatha, swashwin, ztu, ermons, soattos\}@amazon.com}
}

\begin{document}
\maketitle
\input{sec/0_abstract}    
\input{sec/1_intro}
\input{sec/2_related}
\input{sec/3_unet}

\input{sec/4_transformer}

\input{sec/5_data}

\input{sec/6_scaling_law}
\input{sec/7_conclusion}
{
    \small
    \bibliographystyle{ieeenat_fullname}
    \bibliography{main}
}
\appendix
\input{sec/X_suppl}

\end{document}

%% file: preamble.tex
\usepackage[dvipsnames]{xcolor}


%% file: sec/0_abstract.tex
\begin{abstract}
Scaling up model and data size has been quite successful for the evolution of LLMs. However, the scaling law for the diffusion based text-to-image (T2I) models is not fully explored. It is also unclear how to efficiently scale the model for better performance at reduced cost.  The different training settings and expensive training cost make a fair model comparison extremely difficult. In this work, we empirically study the scaling properties of diffusion based T2I models by performing extensive and rigours ablations on scaling both denoising backbones and training set, including training scaled UNet and Transformer variants ranging from 0.4B to 4B parameters on datasets upto 600M images. For model scaling,  we find the location and amount of cross attention distinguishes the performance of existing UNet designs. And increasing the transformer blocks is more parameter-efficient for improving text-image alignment than increasing channel numbers. We then identify an efficient UNet variant, which is 45\% smaller and 28\% faster than SDXL's UNet. On the data scaling side, we show the quality and diversity of the training set matters more than simply dataset size. Increasing caption density and diversity improves text-image alignment performance and the learning efficiency. Finally, we provide scaling functions to predict the text-image alignment performance as functions of the scale of model size, compute and dataset size.

\end{abstract}

%% file: sec/1_intro.tex
\section{Introduction}
\label{sec:intro}

Scaling up model and dataset size has been the key enabling factor for the success of LLMs~\cite{kaplan2020scaling, hoffmann2022training} and VLMs~\cite{clip, cherti2023reproducible}. 
The scaling law~\cite{brown2020language,kaplan2020scaling} governs the expectation of performance  as a function of dataset, model size and compute budget. 
However, the scaling properties for recent diffusion based Text-to-Image (T2I) models \cite{dalle2, ldm, imagen, sdxl} are not well studied. 
Though there is emerging trend that T2I models can be improved with larger denoising backbones~\cite{sdxl, deepfloyd} and stronger text-encoders~\cite{imagen, ediff, sdxl}, it is still not clear how to effectively and efficiently scale up diffusion models, e.g.,
how does the design of denoising backbone influence the image generation and which components are more effective to scale? 
How should diffusion model scale when the training data increases?
To answer the questions, it is essential to understand how exactly each new model improves over previous ones. However, existing diffusion based T2I models are mostly trained with different datasets, input space (latent space or pixel space) and training settings. Moreover, the expensive training cost of high resolution models makes the fair comparison extremely hard, not to mention exploring new ones.
Therefore, a fair and controlled comparison of different denoising backbones is greatly desired,  which can enable seeking of more efficient models with reduced training and inference cost.

\begin{figure}[t]
\centering
\includegraphics[width=0.95\linewidth]{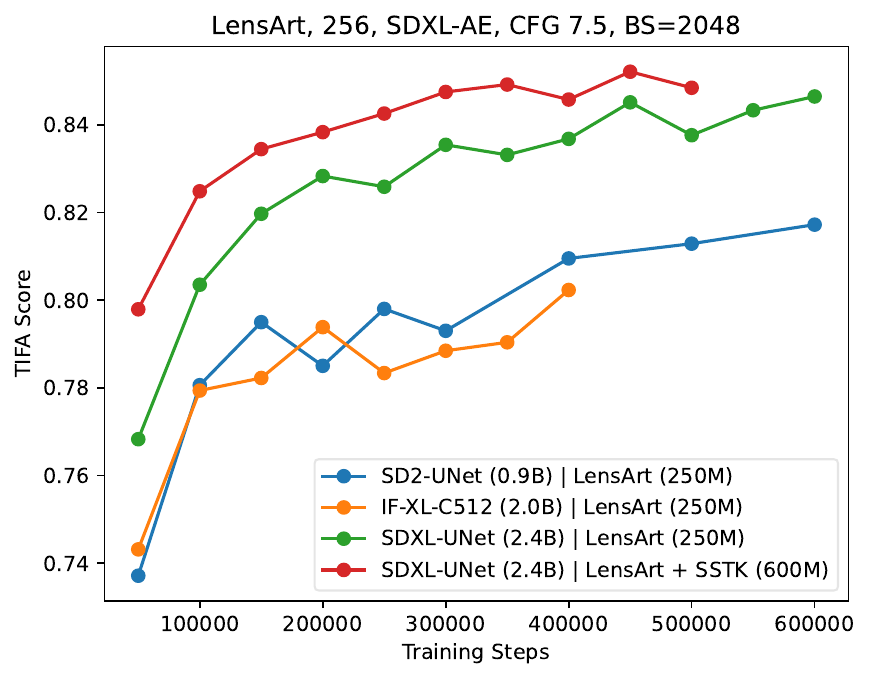}
\caption{Pushing the Pareto frontier of the text-image alignment learning curve by efficiently scaling up both denoising backbones and training data. 
Comparing with the baseline SD2 UNet~\cite{ldm}, the combined scaling with both SDXL UNet and enlarged dataset significantly increases the performance and speeds up the convergence of TIFA score by 6$\times$.}
\label{fig:frontier}
\end{figure}

In this paper, we investigate the scaling properties for training diffusion models, especially on the denoising backbone and dataset.  
The goal is to understand which dimension of the model is more effective and efficient to scale,  how to properly scale the dataset, and the scaling law among models, dataset and compute. Fig.\ref{fig:frontier} gives an illustration of how the Pareto frontier of the text-image alignment performance curve can be pushed via proper scaling.
\subsection{What we have done}
\begin{itemize}
    \item \textbf{Comparing existing UNets in a controlled environment}: 
    we first compare existing UNet designs from SD2~\cite{ldm}, DeepFloyd~\cite{deepfloyd} and SDXL~\cite{sdxl}, to understand why certain UNet design is significantly better than others.
    To allow a fair comparison, we train all models with the same \emph{dataset}, \emph{latent space}, \emph{text encoder} and \textit{training settings}.
    We monitor multiple evaluation metrics during training, including composition scores and image quality scores. 
    We verified SDXL’s UNet achieves superior performance over others with similar amount of parameters, which justifies the importance of architecture design.
    \item \textbf{Scaling UNet and comparing with Transformers}: 
    To understand why SDXL works so well, we conduct extensive ablation studies on the design sapce of UNet by investigating 15 variations ranging from 0.4B to 4B parameters, especially on the choice of channel numbers and transformer depth. 
    We show how each architecture hyperparameter affects the performance and convergence speed. Similarly, we ablate and scale the Transformer backbones~\cite{dit, pixart} and compare with UNet.
    \item \textbf{Ablating the effect of dataset scaling and caption enhancement}: We study how different dataset properties affect the training performance, including dataset size, image quality and caption quality.
    We curate two large-scale datasets with 250M and 350M images, both are augmented by synthetic captions.
    We train both small and large models to see how they can benefit from dataset scaling. 
    
    
\end{itemize}

\subsection{Contributions}
\begin{itemize}
    \item We conduct large-scale controlled experiments to allow fair comparison across various denoising backbones for T2I synthesis, including both UNets and Transformers. 
    Our work verifies the importance of the denoising backbone design. 
    We find composition ability is mainly developed at low resolution, which enables fast model ablations without training in high resolution. 
    To our best knowledge, our work is the first large-scale controlled study allowing fair comparison across different denoising backbones for T2I syntheis.
    \item We ablate the key design factors for UNet and Transformers and compared their scaled versions. 
    We show scalling the transformer depth in UNet is more parameter efficient in improving the alignment performance in comparison with channel number.
    We identify an efficient UNet variant that is 45\% smaller and 28\% faster than SDXL while achieving similar performance.
    We confirm scaling transformer backbone improves performance, but also identify the difficulty of training from scratch due to lack of inductive bias in comparison with UNets.
    \item 
    We show that properly scaling training data with synthetic captions improves image quality and speeds up the convergence. We see data scaling can improve small model's  performance significantly, a better designed model can have a higher performance upper bound.
\end{itemize}


%% file: sec/2_related.tex
\section{Related Work}
\paragraph{Diffusion Models}
Diffusion models~\cite{sohl2015deep,ho2020denoising,nichol2021improved,nichol2021glide, ho2022classifier} synthesize samples via an iterative denoising process and have shown superior performance over GAN~\cite{goodfellow2014generative} based methods for image generation~\cite{dhariwal2021diffusion}. 
Recent diffusion based T2I models such as Imagen~\cite{imagen}, LDM/SD2~\cite{ldm}, DeepFloyd~\cite{deepfloyd}, SDXL~\cite{sdxl}, and DALL$\cdot$E ~\cite{dalle2, dalle3} 
have shown consistently improved performance in terms of sample diversity, text-image alignment and image fidelity. 
Pixel-based models~\cite{dalle2, imagen, deepfloyd} usually require cascaded super-resolution (SR) models to upscale images generated in low resolution, while LDMs~\cite{ldm, sdxl, dalle3} reduce training cost by utilizing a compressed latent space and upsampling with via an autoencoder~\cite{vae}. 
The low resolution latent space may not represent small objects (e.g., faces) well. 
SDXL mitigates this issue via a better VAE and training models in higher resolution latent space (128$\times$128).
Emu~\cite{emu} shows that increasing the latent channels improves image quality.

\paragraph{Scaling UNets}
UNet architecture was first introduced for diffusion models in~\cite{ho2020denoising}.
\cite{nichol2021improved,dhariwal2021diffusion} ablated UNet with several design choices and investigated how FID scales as a function of training compute. 
The UNet in LDM or SD2~\cite{ldm} has 320 initial channels and 850M parameters.
DeepFloyd~\cite{deepfloyd} trains a pixel based model with a UNet of 4B parameter size and 704 channels, which shows better performance than its smaller versions. 
SDXL~\cite{sdxl} employs a 3$\times$ larger UNet than SD2 with mutiple improvements. 
On the other hand, there are also works on improving UNet's efficiency by scaling it down, e.g., 
SnapFusion~\cite{li2023snapfusion} studies the redundancy of UNet and identifies an efficient version by employing the change of CLIP/Latency to measure the impact of architecture change.

\begin{figure*}[t]
\centering
\includegraphics[width=1\linewidth]{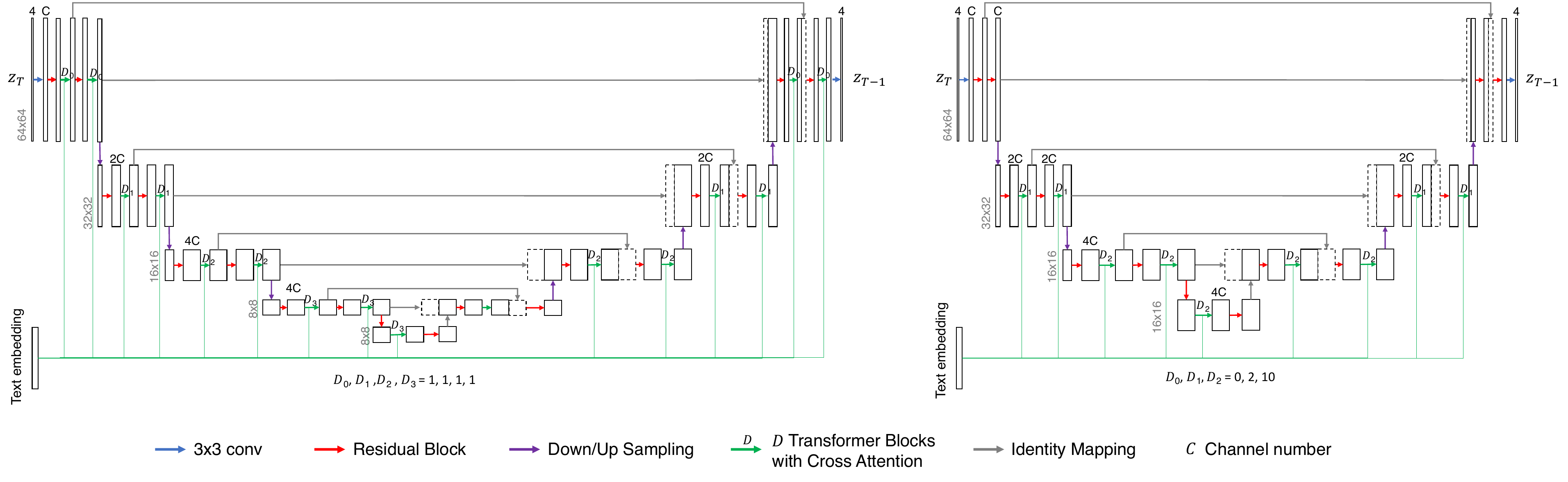}
\caption{Comparison of the UNet design between SD2 (left) and SDXL (right).
SD2 applies cross-attention at all down-sampling levels, including 1$\times$, 2$\times$, 4$\times$ and 8$\times$, while SDXL adopts cross-attention only at 2$\times$ and 4$\times$ down-sampling levels.
}
\label{fig:unet_comparision}
\end{figure*}

\paragraph{Transformer Backbones}
Recently there is surge interest in using Transformer~\cite{vaswani2017attention} to replace UNet for its general architecture design and increased scalability~\cite{dit, bao2023all, zheng2023fast}.
DiT~\cite{dit} replaces UNet with Transformers for class-conditioned image generation and find there is a strong correlation between the network complexity and sample quality.
U-ViT~\cite{bao2023all} shows comparable performance  can be achieved by ViTs with long skip connection. 
MDT~\cite{mdt} introduces a mask latent modeling scheme to improve the training efficiency of transformer-based diffusion models. 
Those works are mostly class conditioned models and only the effect of model architecture on image fidelity is studied.
PixArt-$\alpha$~\cite{pixart} extends DiTs~\cite{dit} for text-conditioned image generation.  More recently, SD3~\cite{sd3} propose MM-DiT design and find it scales well.



%% file: sec/3_unet.tex
\section{Scaling Denoising Backbone}
\label{sec:unet}

\begin{table*}[b]
\center
\scriptsize
\caption{Comparing UNet variants in terms of their hyperparameter, number of parameters, and inference complexity (GMACs). We also list the portion of compute allocated for attention operations.
The original architecture hyperparameters are marked in \textbf{bold}. 
}
\label{tab:design_space}
\begin{tabular}{l|c|c|c|c|c|c|ccc}
\hline
\multirow{2}{*}{UNet}  & \multicolumn{1}{l|}{\multirow{2}{*}{Channels}} & \multicolumn{1}{l|}{\multirow{2}{*}{Channel Mult.}} & \multicolumn{1}{l|}{\multirow{2}{*}{Res. Blocks}} & \multicolumn{1}{l|}{\multirow{2}{*}{Atten. Res.}} & \multicolumn{1}{l|}{\multirow{2}{*}{Tran. Depth}} & \multicolumn{1}{l|}{\multirow{2}{*}{Params (B)}} & \multicolumn{3}{c}{GMACs}                                                              \\ \cline{8-10} 
                      & \multicolumn{1}{l|}{}                          & \multicolumn{1}{l|}{}                               & \multicolumn{1}{l|}{}                             & \multicolumn{1}{l|}{}                             & \multicolumn{1}{l|}{}                    & \multicolumn{1}{l|}{}                            & \multicolumn{1}{l|}{Total}        & \multicolumn{1}{l|}{Atten.}   & \multicolumn{1}{l}{Atten.\%} \\ \hline
\multirow{2}{*}{SD2~\cite{ldm}}   & \textbf{320}                                   & \multirow{2}{*}{\textbf{{[}1,2,4,4{]}}}             & \multirow{2}{*}{\textbf{2}}                       & \multirow{2}{*}{\textbf{{[} 4, 2, 1 {]}}}         & \multirow{2}{*}{\textbf{{[}1, 1, 1{]}}}        & \textbf{0.87}                                    & \multicolumn{1}{c|}{\textbf{86}}  & \multicolumn{1}{c|}{34}  & \textbf{39}           \\ \cline{2-2} \cline{7-10} 
                      & 512                                            &                                                     &                                                   &                                                   &                                          & 2.19                                             & \multicolumn{1}{c|}{219}          & \multicolumn{1}{c|}{85}  & {39}           \\ \hline
\multirow{2}{*}{IF-XL~\cite{deepfloyd}} & 512                                            & \multirow{2}{*}{\textbf{{[}1,2,3,4{]}}}             & \multirow{2}{*}{\textbf{3}}                       & \multirow{2}{*}{\textbf{{[}4, 2, 1{]}}}           & \multirow{2}{*}{\textbf{{[}1, 1, 1{]}}}        & 2.04                                             & \multicolumn{1}{c|}{194}          & \multicolumn{1}{c|}{23}  & 12                    \\ \cline{2-2} \cline{7-10} 
                      & \textbf{704}                                   &                                                     &                                                   &                                                   &                                          & 3.83                                             & \multicolumn{1}{c|}{364}          & \multicolumn{1}{c|}{42}  & \textbf{12}                    \\ \hline
                      
\multirow{12}{*}{SDXL~\cite{sdxl} } & 128                                            & \multirow{4}{*}{\textbf{{[}1, 2, 4{]}}}             & \multirow{4}{*}{\textbf{2}}                       & \multirow{4}{*}{\textbf{{[}4, 2{]}}}              & \multirow{4}{*}{\textbf{{[}0, 2, 10{]}}} & 0.42                                             & \multicolumn{1}{c|}{35}           & \multicolumn{1}{c|}{23}  & 65                   \\ \cline{2-2} \cline{7-10} 
                      & 192                                            &                                                     &                                                   &                                                   &                                          & 0.90                                             & \multicolumn{1}{c|}{75}           & \multicolumn{1}{c|}{48}  & 65                    \\ \cline{2-2} \cline{7-10} 
                      & \textbf{320}                                   &                                                     &                                                   &                                                   &                                          & \textbf{2.39}                                    & \multicolumn{1}{c|}{\textbf{198}} & \multicolumn{1}{c|}{127} & \textbf{64}           \\ \cline{2-2} \cline{7-10} 
                      & 384                                            &                                                     &                                                   &                                                   &                                          & 3.40                                             & \multicolumn{1}{c|}{282}          & \multicolumn{1}{c|}{179} & 64                    \\ \cline{2-10} 
                      & \multirow{7}{*}{\textbf{320}}                  & \multirow{7}{*}{\textbf{{[}1, 2, 4{]}}}             & \multirow{7}{*}{\textbf{2}}                       & \multirow{7}{*}{\textbf{{[}4, 2{]}}}              & {[}0, 2, 2{]}                            & 0.85                                             & \multicolumn{1}{c|}{98}           & \multicolumn{1}{c|}{43}  & 44                    \\ \cline{6-10} 
                      &                                                &                                                     &                                                   &                                                   & {[}0, 2, 4{]}                            & 1.24                                             & \multicolumn{1}{c|}{123}          & \multicolumn{1}{c|}{64}  & 52                    \\ \cline{6-10} 
                      &                                                &                                                     &                                                   &                                                   & {[}0, 2, 12{]}                           & 2.78                                             & \multicolumn{1}{c|}{223}          & \multicolumn{1}{c|}{147} & 66                    \\ \cline{6-10} 
                      &                                                &                                                     &                                                   &                                                   & {[}0, 2, 14{]}                           & 3.16                                             & \multicolumn{1}{c|}{248}          & \multicolumn{1}{c|}{168} & 68                   \\ \cline{6-10} 
                      &                                                &                                                     &                                                   &                                                   & {[}0, 4, 4{]}                            & 1.32                                             & \multicolumn{1}{c|}{143}          & \multicolumn{1}{c|}{84}  & {59}           \\ \cline{6-10} 
                      &                                                &                                                     &                                                   &                                                   & {[}0, 4, 8{]}                            & 2.09                                             & \multicolumn{1}{c|}{193}          & \multicolumn{1}{c|}{123} & 64                    \\ \cline{6-10} 
                      &                                                &                                                     &                                                   &                                                   & {[}0, 4, 12{]}                           & 2.86                                             & \multicolumn{1}{c|}{243}          & \multicolumn{1}{c|}{167} & 69                    \\ \cline{2-10} 
                      & 384                                            & \textbf{{[}1, 2, 4{]}}                              & \textbf{2}                                        & \textbf{{[}4, 2{]}}                               & {[}0, 4, 12{]}                           & 4.07                                             & \multicolumn{1}{c|}{346}          & \multicolumn{1}{c|}{237} & 69                    \\
\hline
\end{tabular}
\end{table*}

\subsection{Existing UNet Design}
The UNet in diffusion models adopts a stack of residual blocks and a sequence of downsampling and upsampling convolutions, along with additional spatial attention layers at multiple resolutions~\cite{ho2020denoising,dhariwal2021diffusion}.
Recent T2I frameworks~\cite{ldm,sdxl,deepfloyd} mostly employ the ideas in simple diffusion~\cite{hoogeboom2023simple} to improve the efficiency of UNet, i.e., tweaking more parameters and computation at smaller resolutions.
Fig.~\ref{fig:unet_comparision} gives a comparison of the UNets for SD2 and SDXL. SDXL improves over SD2 in multiple dimensions:
a) \textbf{Less downsampling rates.} SD2 uses [1, 2, 4, 4] as the multiplication rates to increase channels at different downsampling levels. 
DeepFloyd adopts [1, 2, 3, 4] to reduce computation, while SDXL uses [1, 2, 4], which completely removes the 4th downsampling level.
b) \textbf{Cross-attention only at lower resolution.} Cross-attention is only computed at certain downsampling rates, e.g., SD2 applies cross-attention at first three downsampling rates (1$\times$, 2$\times$, 4$\times$), while SDXL only integrates text embedding at the 2$\times$ and 4$\times$ downsampling levels.
c) \textbf{More compute at lower resolution.} SDXL applies more transformer blocks at the 2$\times$ and 4$\times$ downsampling levels, while SD2 applies uniform single transformer block at all three downsampling levels.

\subsection{Controlled Comparison of UNets}
To allow fair comparison of different UNets, we train all backbone variants in the same controlled settings, including the same \emph{dataset}, \emph{latent space}, \emph{text-encoder} and \textit{training settings}.
Below we introduce the training configurations and evaluation metrics, based on which we compare all different backbones variants.

\paragraph{Training} 
We train models on our curated dataset \emph{LensArt}, which contains 250M text-image pairs (details in Sec~\ref{sec:dataset}). 
We use SDXL's VAE and the OpenCLIP-H~\cite{openclip} text encoder (1024 dim), without adding extra embedding layer or other conditioning.
We train all models at 256$\times$256 resolution with batch size 2048 upto 600K steps.
We follow the setup of LDM~\cite{ldm} for DDPM schedules.
We use AdamW~\cite{adamw} optimizer with 10K steps warmup and then constant learning rate 8e-5.
We employ mixed precision training with BF16 and enables FSDP for large models.

\paragraph{Inference and Evaluation} 
We use DDIM sampler~\cite{song2020denoising} in 50 steps with fixed seed and CFG scale (7.5) for inference.
To understand the training dynamics, we monitor the evolution of five metrics during training. 
We find the the metrics at early stage of training can help predict final model performance.
Specifically, we measure composition ability and image quality with metrics including: 
1) \textbf{TIFA}~\cite{tifa}, which measures the faithfulness of a generated image to its text input via visual question answering (VQA). 
It contains 4K collected prompts and corresponding  question-answer pairs generated by a language model. Image faithfulness is calculated by checking whether existing VQA models can answer these questions using the generated image. TIFA allows for fine-grained and interpretable evaluations of generated images.  
2) \textbf{ImageReward}~\cite{imagereward} which was learned to approximates human preference. We calculate the average ImageReward score over images generated with MSCOCO-10K prompts. Though ImageReward is not a normalized score, its scores are in the range of [-2, 2] and the average score over the number of images gives meaningful statistics to allow comparision across models.
Due to space constraints, we mainly show TIFA and ImageReward and provide results of other metrics (CLIP score~\cite{clip, clipscore}, FID, HPSv2~\cite{hpsv2}) in Appendix.

\begin{figure}[t]
\centering
\includegraphics[width=0.495\linewidth]{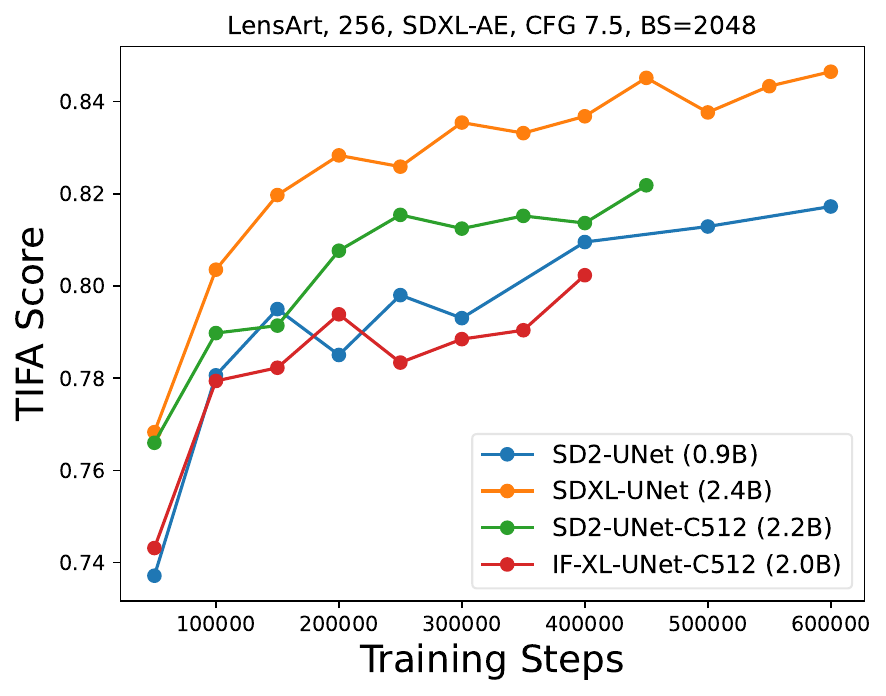}
\includegraphics[width=0.495\linewidth]{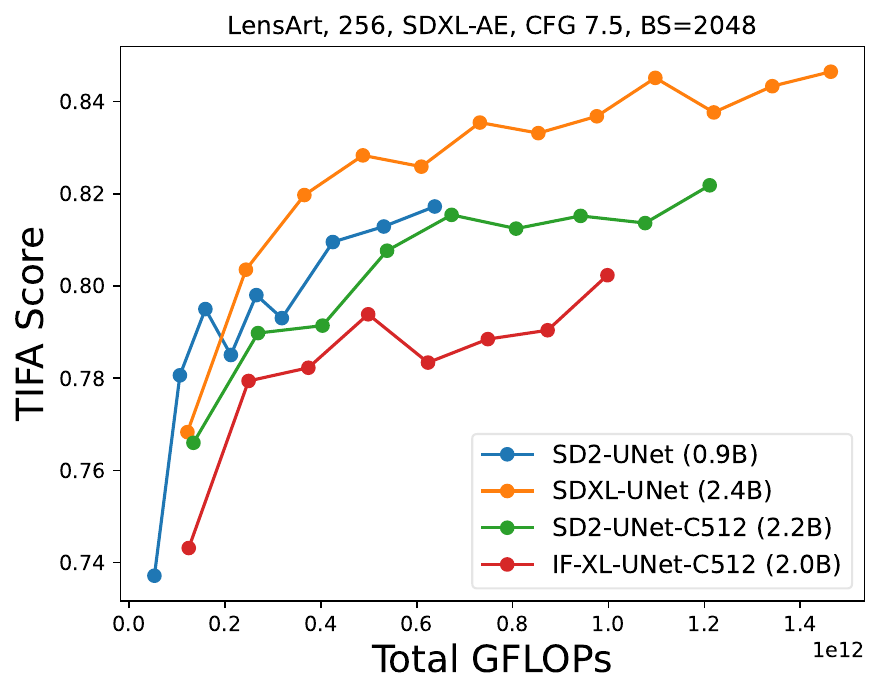}
\caption{The evolution of TIFA score during training with different UNets on the same dataset in terms of training steps and training compute (GFLOPs). 
The compute FLOPs is estimated with 3$\times$ FLOPs of single DDPM step $\times$ batch size $\times$ steps.
}
\vspace{-4mm}
\label{fig:sdxl_vs_others}
\end{figure}

\paragraph{SDXL vs SD2 vs IF-XL} 
We compare the design of several existing UNet models from SDXL~\cite{sdxl}, DeepFloyd-IF~\cite{deepfloyd},  
SD2~\cite{ldm} and its scaled version in the above controlled settings. 
Specifically, we compare a) 
SD2 UNet (0.9B) b) SD2 UNet with 512 initial channels (2.2B) c) SDXL's UNet (2.4B) d) DeepFloyd's IF-XL UNet with 512 channels (2.0B).
Fig.~\ref{fig:sdxl_vs_others} shows that the naively scaled SD2-UNet (C512, 2.2B) achieves better TIFA score than the base SD2 model at the same training steps. 
However, the convergence speed is slower in terms of training FLOPs, which indicates \emph{increasing channels is an effective but not an efficient approach}.
SDXL's UNet achieves 0.82 TIFA within 150K steps, which is 6$\times$ faster than SD2 UNet and 3$\times$ faster than SD2-C512 in training iterations.
Though its training iteration speed (FLOPS) is 2$\times$ slower than SD2, it still achieves the same TIFA score at 2$\times$ reduced training cost. 
SDXL UNet also can get a much higher TIFA score (0.84) with a clear margin over other models.
Therefore SDXL's UNet design is \emph{significantly better than others in terms of performance and training efficiency, pushing the Pareto frontier}.

\begin{figure*}[t]
\centering
\includegraphics[width=0.33\linewidth]{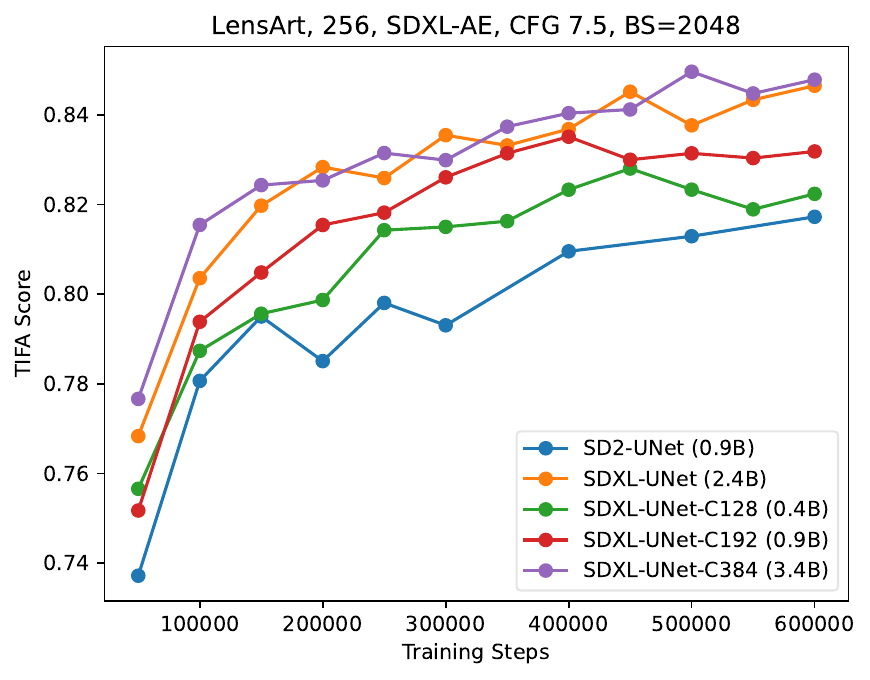}
\includegraphics[width=0.33\linewidth]{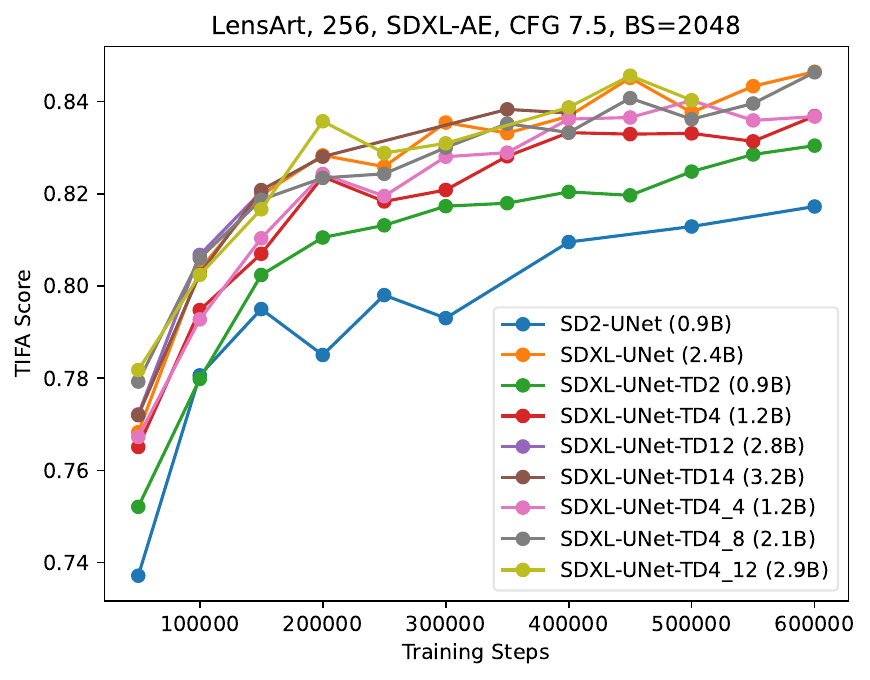}
\includegraphics[width=0.33\linewidth]{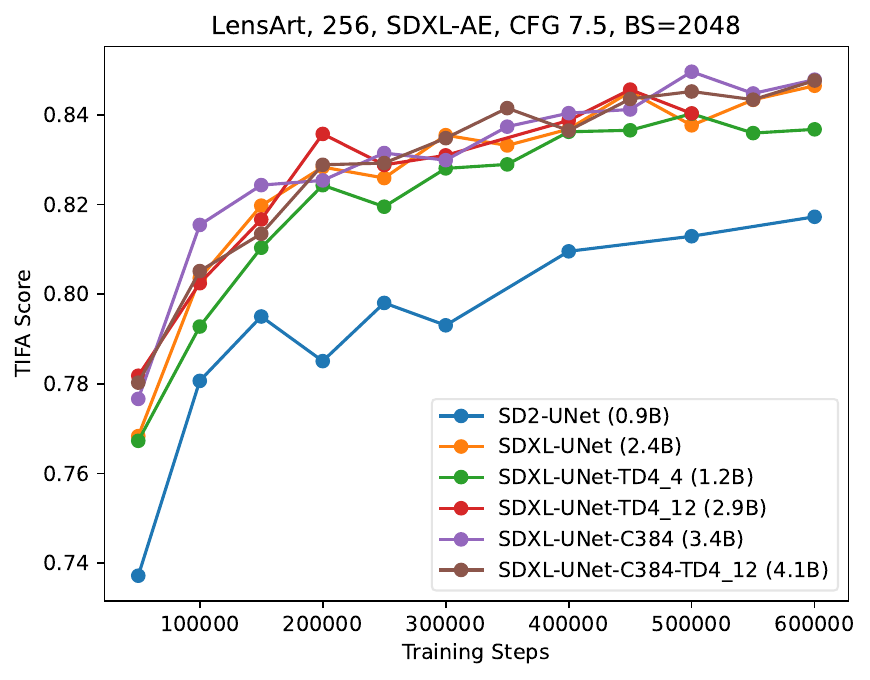}
\caption{Evolution of TIFA score during training with scaled UNet variations. 
The baseline models are UNets of SD2 and SDXL. We train SDXL UNet variants with changes in (a) channels $C$ (b) transformer depth (TD) 3) both channels and TD.}
\label{fig:channels_tds_search}
\end{figure*}

\subsection{Ablation of UNet Design}
Now we have verified SDXL has a much better UNet design than SD2 and DeepFloyd variants.
The question is why it excels and how to further improve it effectively and efficiently.
Here we investigate how to improve SDXL's UNet by exploring its design space.

\paragraph{Search Space} Table~\ref{tab:design_space} shows different UNet configurations, and their corresponding compute complexity at 256 resolution.
We mainly vary the initial channels and transformer depth.
To understand the impact of each dimension of the design space, we select a subset of the variant models and train them with the same configurations.
This forms our main ``search space” for the UNet architecture. 
More ablations on the impact of VAE, training iterations and batch size can be found in Appendix.

\paragraph{The Effect of Initial Channels}
We train the following SDXL UNet variants with different channel numbers: 128, 192, and 384, with parameters 0.4B, 0.9B and 3.4B, respectively.
Fig.~\ref{fig:channels_tds_search} (a) shows that UNet with reduced channels from 320 to 128 still can outperform SD2’s UNet with 320 channels, which shows that \emph{less channel number can achieve better quality with proper architecture design}. 
However, the TIFA (also ImageReward/CLIP) scores are worse in comparison with SDXL’s UNet, which indicates its importance in visual quality. 
Increasing channel number from 320 to 384 boosts the number of parameters from 2.4B to 3.4B. 
It also achieves better metrics than baseline 320 channels at 600K training steps.
Note that the initial channel number $C$ actually connects with other hyperparameters of the UNet, e.g., 
1) the dimension of timestep embedding $T$ is $4C$; 
2) the number of attention head is linear with channel numbers, i.e. $C/4$.
As shown in Table~\ref{tab:design_space}, the proportion of compute for attention layers are stable (64\%) when $C$ changes.
This explains why increasing the width of UNet also brings alignment improvement as shown in Fig.~\ref{fig:channels_tds_search}.

\paragraph{The Effect of Transformer Depth}
The transformer depth (TD) setting controls the number of transformer blocks at certain input resolution. 
SDXL applylies 2 and 10 transformer blocks at 2$\times$ and 4$\times$ downsampling level, respectively. 
To understand its effect, we trained the variants shown in Table~\ref{tab:design_space} with different TDs, ranging from 0.9B to 3.2B parameters.
Specifically, we first change the TD at the 4$\times$ downsampling rate, obtaining TD2, TD4, TD12 and TD14, then we further change the depth at 2$\times$ downsmapling rate, resulting in TD4\_4, TD4\_8 and TD4\_12. 
Note the portion of attention operations also increases with TDs accordingly.
Fig.~\ref{fig:channels_tds_search} (b) shows that increasing TD at 4$\times$ downsampling rate from 2 to 14 continuously improves TIFA score.  
From the comparison between TD4 and TD4\_4, we see that increasing transformer depth at 2$\times$ resolution (2 $\xrightarrow{}$ 4) also improves TIFA score. 
TD4\_4 has competitive performance in comparison with SDXL’s UNet while having 45\% less parameters and 28\% less compute for inference. 
In Appendix, we show TD4\_4 achieves same TIFA score 1.7$\times$ faster than SDXL UNet in terms of wall-clock training time.
TD4\_8 has almost the same performance as SDXL's UNet with 13\% less parameters.
Since the text-image alignment (TIFA) is mostly about large objects in the image, it is helpful to \textit{allocate more cross compute at lower resolution or global image level beyond efficiency considerations}.

\paragraph{Scaling both Channels and Transformer Depth}
Given the effect of channels and transformer depth, we further explored enlarging both the channel numbers (from 320 to 384) and transformer depth ([0, 2, 10] $\xrightarrow{}$ [0, 4, 12]). 
Fig.~\ref{fig:channels_tds_search} (c) shows that it achieves slightly higher TIFA scores during training than SDXL-UNet.
However, the advantage over simply increasing channels or transformer depth is not apparent,
which means \textit{there is a performance limit for models to continue scale under metrics like TIFA}.

\begin{figure}[t]
\centering
\includegraphics[height=0.46\linewidth]{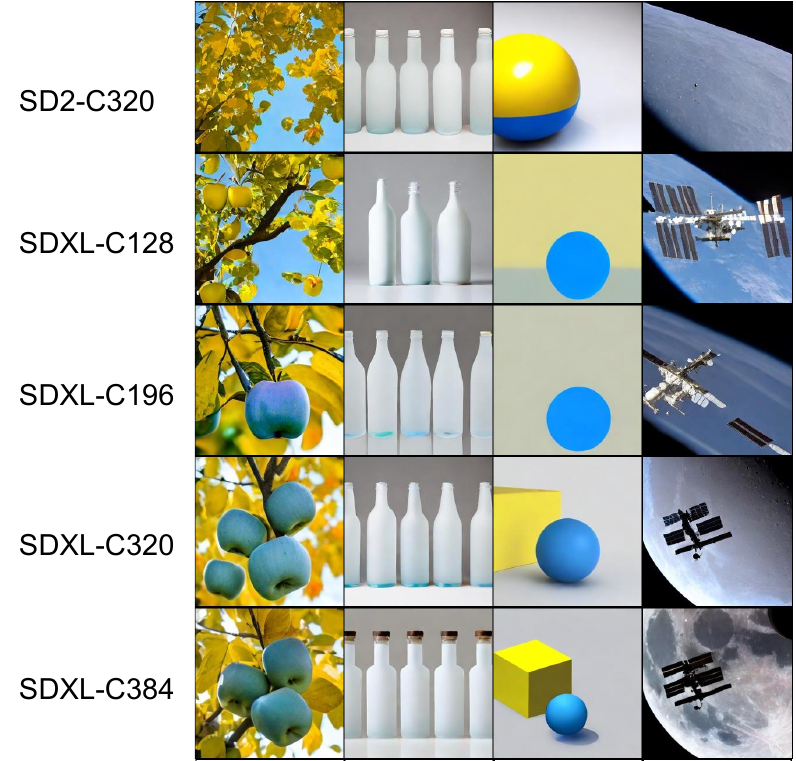}
\includegraphics[height=0.46\linewidth]{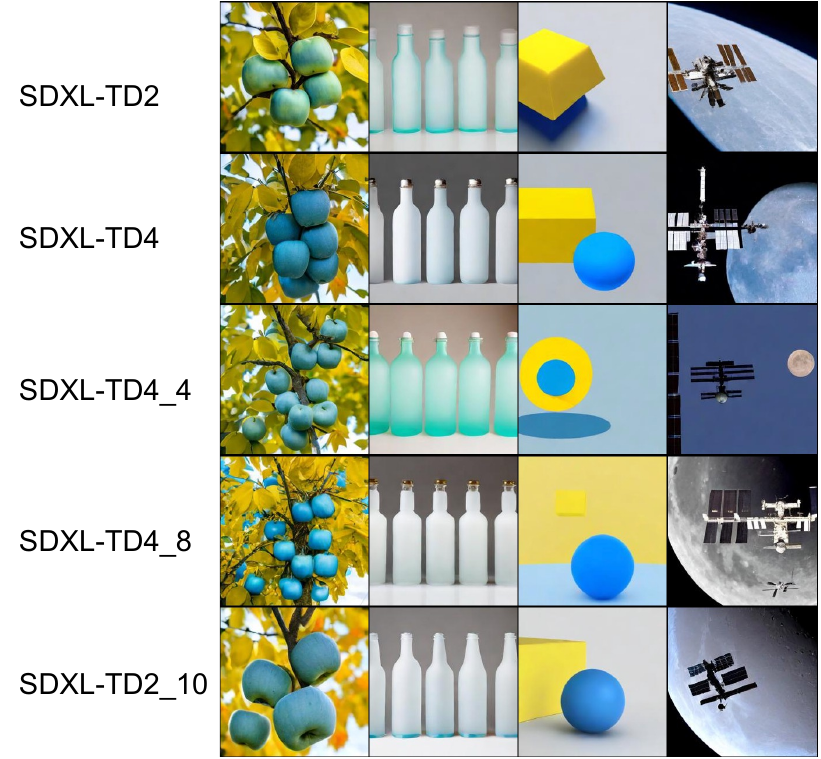}

\caption{
Visualizing the effect of UNet scaling on text-image alignment.
We change the UNet along two dimensions: channel number (left) and transformer depth (right).
The prompts are: 1) "square blue apples on a tree with circular yellow leaves" 2) "five frosted glass bottles" 3) "a yellow box to the right of a blue sphere" 4) "the International Space Station flying in front of the moon"}
\label{fig:visual}
\end{figure}

\paragraph{Visualizing the Effect of UNet Scaling}
Fig.~\ref{fig:visual} shows the images generated by different UNets with the same prompts.
We can see that as the channel number or transformer depth increases, the images become more aligned with the given prompts (e.g., color, counting, spatial, object). 
Some images generated by certain UNet variant are better than the original SDXL UNet (C320), i.e., SDXL-C384 and SDXL-TD4\_8 both generate more accurate images with the 4th prompt.

%% file: sec/4_transformer.tex
\subsection{Comparing with Transformers}
\label{sec:transformer}

\begin{table*}[tb]
\scriptsize
\caption{Hyperparameter settings for Transformer-based backbones at resolution 256x256.
The original PixArt-$\alpha$~\cite{pixart} model uses T5-XXL tokenizer, while we use OpenCLIP-H to keep consistency with UNet experiments.
The original architecture settings are marked in \textbf{bold}. 
$p$, $h$ and $d$ denote patch size, hidden dimension, and depth.
}
\label{tab:transformer}
\centering
\begin{tabular}{l|c|c|c|c|c|c|c|c|c|c|c}
\hline
\multirow{2}{*}{Model}                           & \multicolumn{1}{l|}{\multirow{2}{*}{VAE}} & \multicolumn{1}{l|}{\multirow{2}{*}{$p$}} & \multicolumn{1}{l|}{\multirow{2}{*}{$h$}} & \multicolumn{1}{l|}{\multirow{2}{*}{$d$}} & \multicolumn{1}{l|}{\multirow{2}{*}{\#heads}} & \multicolumn{1}{l|}{\multirow{2}{*}{Text Encoder}} & \multicolumn{1}{l|}{\multirow{2}{*}{Max Tokens}} & \multicolumn{1}{l|}{\multirow{2}{*}{Token Dim.}} & \multicolumn{1}{l|}{\multirow{2}{*}{Cap. Emb}} & \multicolumn{1}{l|}{\multirow{2}{*}{GMACs}} & \multicolumn{1}{l}{\multirow{2}{*}{Params (B)}} \\
                                                 & \multicolumn{1}{l|}{}                     & \multicolumn{1}{l|}{}                     & \multicolumn{1}{l|}{}                     & \multicolumn{1}{l|}{}                     & \multicolumn{1}{l|}{}                         & \multicolumn{1}{l|}{}                              & \multicolumn{1}{l|}{}                            & \multicolumn{1}{l|}{}                           & \multicolumn{1}{l|}{}                              & \multicolumn{1}{l|}{}                       & \multicolumn{1}{l}{}                            \\ \hline
{PixArt-$\alpha$-XL/2\cite{pixart}} & \textbf{SD2}                              & \textbf{2}                                & \textbf{1152}                             & \textbf{28}                               & \textbf{16}                                   & \textbf{4.3B Flan-T5-XXL}                          & \textbf{120}                                     & \textbf{4096}                                   & \textbf{Y}                                         & 139                                         & 0.61                                             \\ \hline
\multirow{4}{*}{Ours}                            & \multirow{4}{*}{SDXL}                     & \multirow{4}{*}{\textbf{2}}               & \multirow{1}{*}{\textbf{1152}}            & \multirow{2}{*}{\textbf{28}}                                 & \multirow{2}{*}{\textbf{16}}                                   & \multirow{4}{*}{354M OpenCLIP-H}                   & \multirow{4}{*}{77}                              & \multirow{4}{*}{1024}                           & \textbf{Y}                                         & 139                                         & 0.61 \\ 
\cline{4-4}  \cline{10-12} 
   &                     &                & {1536}            &                                &                                   &                  &                               &                            & \textbf{Y}                                         & 247                                         &  1.08\\ 
\cline{4-6} \cline{10-12}  
                                                 &                                           &                                           & \multirow{2}{*}{1024}                     & \textbf{28}                               & \multirow{2}{*}{\textbf{16}}                                    &                                                    &                                                  &                                                 & N                                                  & 110                                         & 0.48 \\ \cline{5-5} \cline{10-12} 
&                                           &                                           &                                           & 56                                        &                                    &                                                    &                                                  &                                                 & N                                                  & 220                                         & 0.95                                             \\ \hline
\end{tabular}
\end{table*}

DiT~\cite{dit} demonstrates that scaling up the transformer complexity can get consistently improved image fidelity for class-conditioned image generation on ImageNet.
PixArt-$\alpha$~\cite{pixart} extends DiT to text-conditioned image generation with similar backbone. 
However, there is a lack of fair comparison with UNet in a controlled setting.
To compare with UNet and understand its scaling property, we train multiple scaled version of PixArt-$\alpha$, keeping other components and settings the same as previous ablations. 
Table~\ref{tab:transformer} shows the configuration of our scaled variants.
The difference with the original PixArt-$\alpha$ model lies at: 
1) we use SDXL’s VAE instead of SD2’s VAE; 
2) we use OpenCLIP-H text encoder instead of T5-XXL~\cite{t5}, the token embedding dimension is reduced from 4096 to 1024. The token length is 77 instead of 120. 

\begin{figure}[t]
\centering
\includegraphics[width=0.49\linewidth]{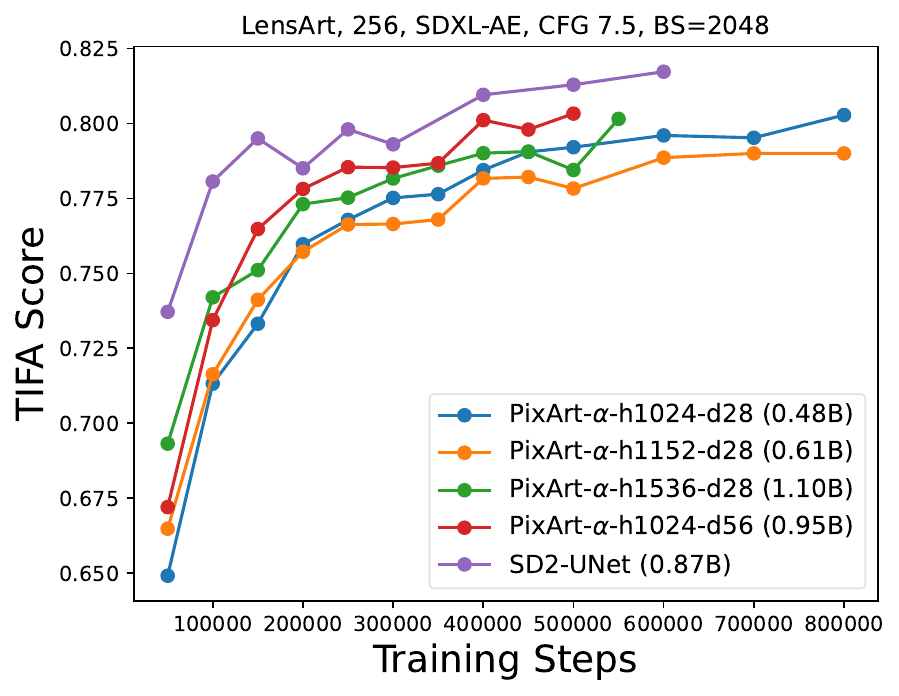}
\includegraphics[width=0.49\linewidth]{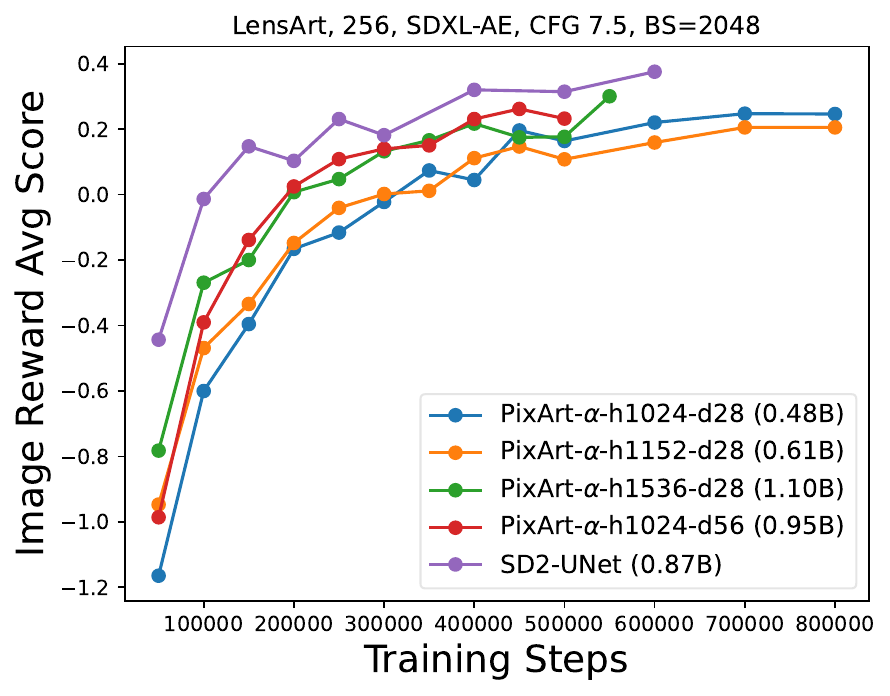}
\caption{The evolution of TIFA and ImageReward during training for scaled PixArt-$\alpha$ variants as well as SD2 UNet.}
\vspace{-4mm}
\label{fig:transformer}
\end{figure}

\paragraph{Ablation Space} We ablate the PixArt-$\alpha$ model in the following dimensions:
1) \emph{hidden dimension $h$}: PixArt-$\alpha$ inherits the design of DiT-XL/2~\cite{dit}, which has 1152 dimension. We also consider 1024 and 1536. 
2) \emph{transformer depth $d$}: we scale the transformer depth from 28 to 56.
3) \emph{caption embedding}: the caption embedding layer maps the text encoder output to dimension $h$.
When the hidden dimension is the same as text embedding (i.e., 1024), we can skip the caption embedding to use the token embedding directly.

\paragraph{The Effect of Model Scaling} 
As shown in Fig.~\ref{fig:transformer}, scaling the hidden dimension $h$ and model depth $d$ both result in improved text-image alignment and image fidelity, while scaling depth $d$ changes model's compute and size linearly. 
Both $d$56 and $h1536$ variants achieve \~1.5$\times$ faster convergence speed than the baseline $d$28 model with similar parameter size and compute.

\paragraph{Comparison with UNet}
The PixArt-$\alpha$ variants yield lower TIFA and ImageReward scores in comparison with SD2-UNet trained in same steps, e.g., SD2 UNet reaches 0.80 TIFA and 0.2 ImageReward at 250K steps while the 0.9B PixArt-$\alpha$ variant gets 0.78 and 0.1.
PixArt-$\alpha$ \cite{pixart} also reports that training without ImageNet pre-training tends to generate distorted images in comparison to models initialized from pre-trained DiT weights, which is trained 7M steps on ImageNet~\cite{dit}.
Though DiT~\cite{dit} proves the UNet is not a must for diffusion models, the PixArt-$\alpha$ variants take longer iterations and more compute to achieve similar performance as UNet.
We leave this improvement for future works and expect architecture improvement can mitigate this issue, such as works done in~\cite{mdt,sd3, yan2023diffusion}.

%% file: sec/5_data.tex
\section{Scaling Training Data}
\label{sec:dataset}

\subsection{Dataset Curation}
We curate our own datasets named \emph{LensArt} and \emph{SSTK}.
\emph{LensArt} is a 250M image-text pairs sourced from 1B noisy web image-text pairs. 
We apply a series of automatic filters to remove data noise, including but not limited to NSFW content, low aesthetic images, duplicated images, and small images.
\emph{SSTK} is another internal dataset with about 350M cleaned data. Table~\ref{tab:dataset} shows the statistics of the datasets. 
More detailed analysis can be seen in the Appendix.

\begin{table}[h!]
\center
\scriptsize
\caption{Dataset statistics. 
\textbf{I-C} indicates the number of unique image-caption pairs. \textbf{AE} indicates the average aesthetic score for the dataset. 
\textbf{I-N} indicates the total number of image-noun pairs, where each pair includes the image and a noun that is unique within corresponding real and synthetic captions.
\textbf{UN} indicates the number of unique nouns within whole text corpus (here nouns includes both nouns and proper nouns for simplicity). 
\textbf{N/I} indicates the average number of nouns per image.
\textbf{w. Syn} indicates the dataset incorporates synthetic captions.}
\label{tab:dataset}
\begin{tabular}{l|c|l|c|l|c}\hline
\textbf{Datasets}           & \textbf{I} & \textbf{AE} & \textbf{I-N} & \textbf{UN}                        & \textbf{N/I}     \  \\ \hline
LensArt-raw                                                   & 1.0B   &             5.00                     &      7.1B                              &           3.9M                         & 7.1                         \\ \hline
LensArt                    & 250M   & {5.33} & {1.8B} & {1.2M} & {7.0}  \\ \hline
SSTK        & 360M   &   5.20                               &                    2.2B                &                           680K         &          6.0                  \\ \hline
LensArt + SSTK                    & 610M   &      5.25                            &    3.9B                                &              1.7M                      &            6.5               \\ \hline
LensArt (w. Syn) & 250M   & {5.33} &  3.2B                                  &                     1.3M               &  12.8                        \\ \hline
SSTK (w. Syn)        & 360M   &   5.20                               &                    4.2B                &                           1.1M         &          11.6                  \\ \hline
LensArt + SSTK (w. Syn)                    & 610M   &     5.25                             &        7.3B                            &                2.0M                    &      12.2                    \\ \hline
\end{tabular}
\end{table}

\begin{figure*}[t]
\begin{minipage}[t]{.48\textwidth}
\includegraphics[width=0.48\linewidth]{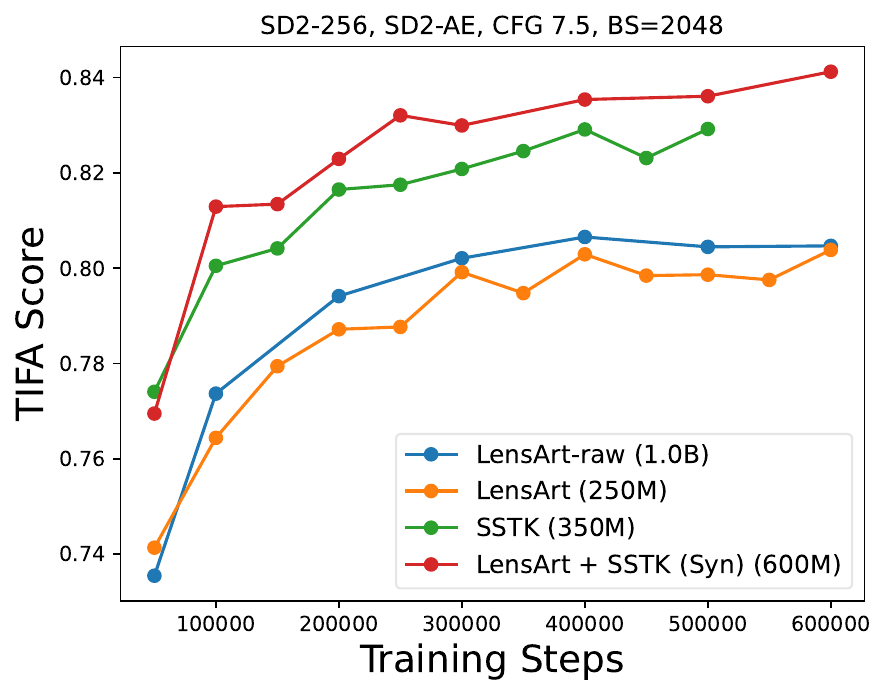}
\includegraphics[width=0.48\linewidth]{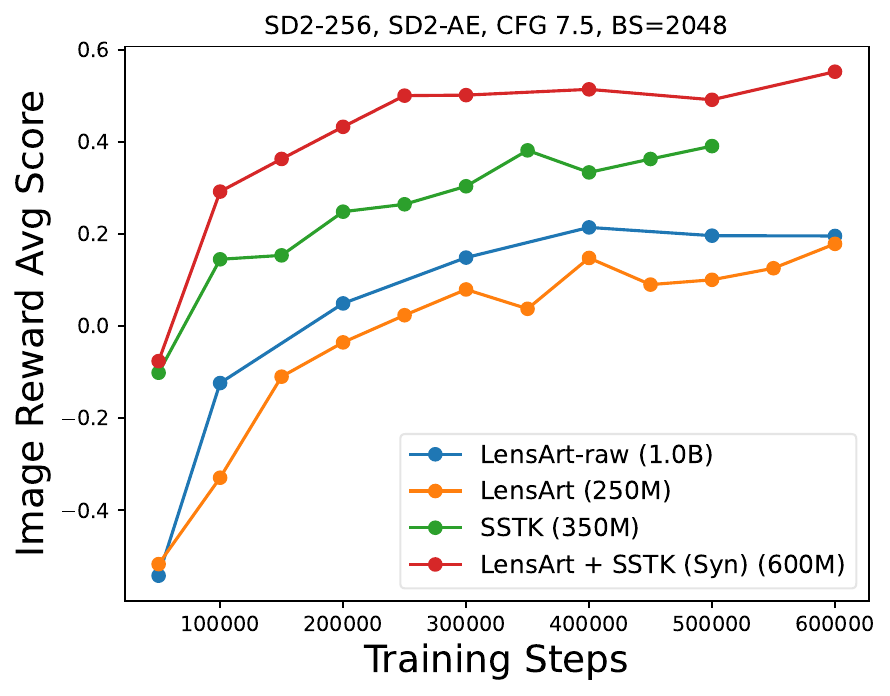}
\caption{
\footnotesize
SD2 models trained on different datasets and their corresponding TIFA and average ImageReward scores during training.
Increasing the scale of the dataset by combining LensArt and SSTK gets the best results.
}
\label{fig:datasetcurves}
\end{minipage}
\hspace{6mm}
\begin{minipage}[t]{.48\textwidth}
\includegraphics[width=0.48\linewidth]{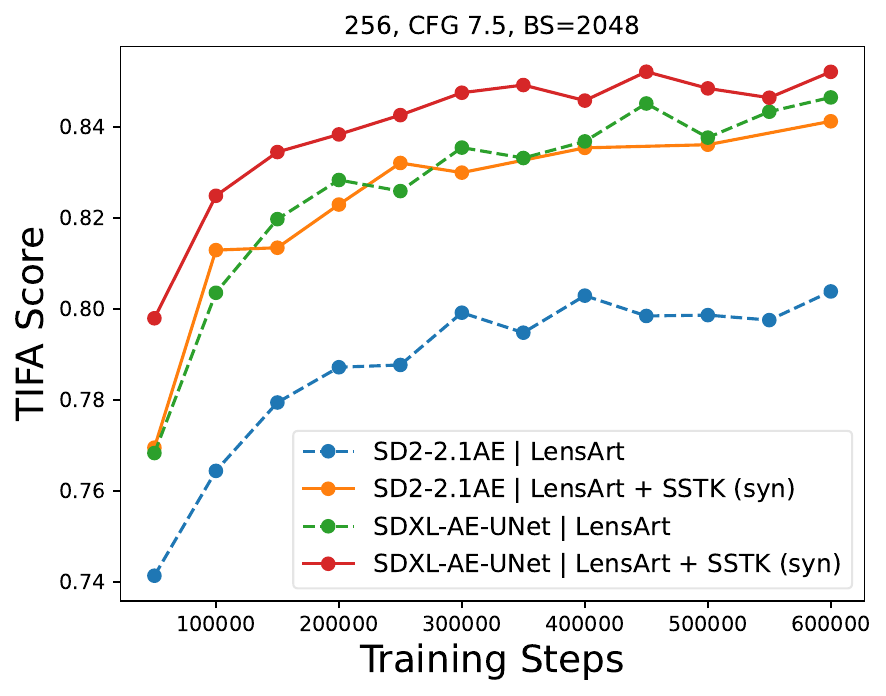}
\includegraphics[width=0.48\linewidth]{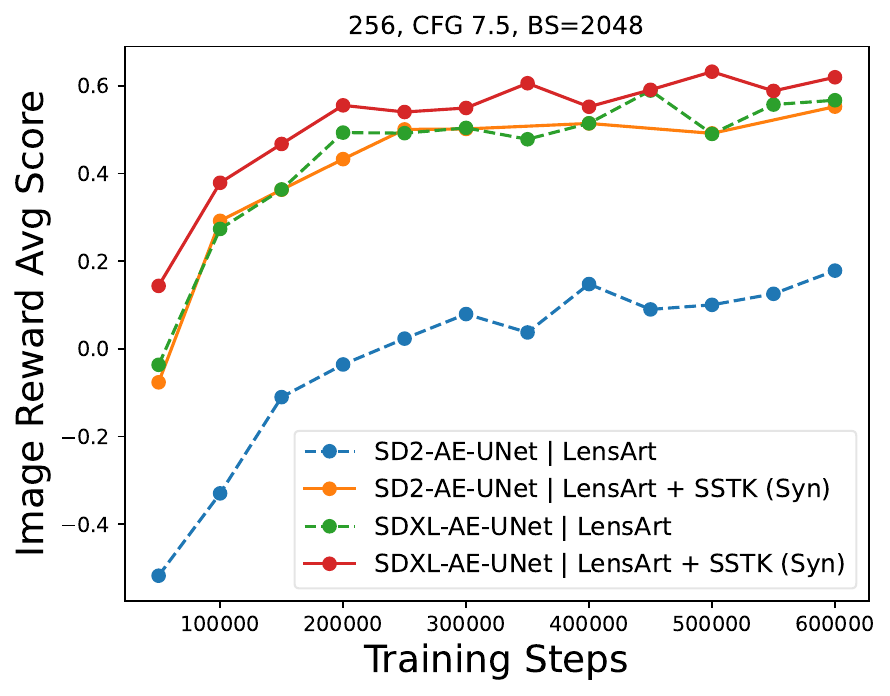}
\caption{
\footnotesize
Comparing SD2 and SDXL trained on LensArt and LensArt+SSTK (w.Syn). Enlarging training set helps improving model performance. Stronger model yields better performance with larger dataset.}
\label{fig:scale_model_dataset}
\end{minipage}
\end{figure*}

\subsection{Data Cleaning} 
The quality of the training data is the prerequisite of data scaling. Compared to training with noisy data source, a high-quality subset not only improves image generation quality, but also preserves the image-text alignment.
LensArt is 4$\times$ smaller than its unfiltered 1B data source, with hundreds of million noisy data removed. 
However, model trained with this high-quality subset improves the average aesthetic score \cite{laion_aev2} on the generated images from 5.07 to 5.20. This is caused by LensArt having an average aesthetic score of 5.33, higher than 5.00 in LensArt-raw. Moreover, as shown in Fig.~\ref{fig:datasetcurves}, a SD2 model trained in LensArt achieves similar TIFA score in comparison the one trained with the raw version, demonstrating that the filtering does not hurt image-text alignment. 
The reason is that sufficient commonsense knowledge is still retained under aggressive filtering while enormous duplicated and long-tail data removed. 

\begin{figure}[t!]
\centering
\includegraphics[width=1.0\linewidth]{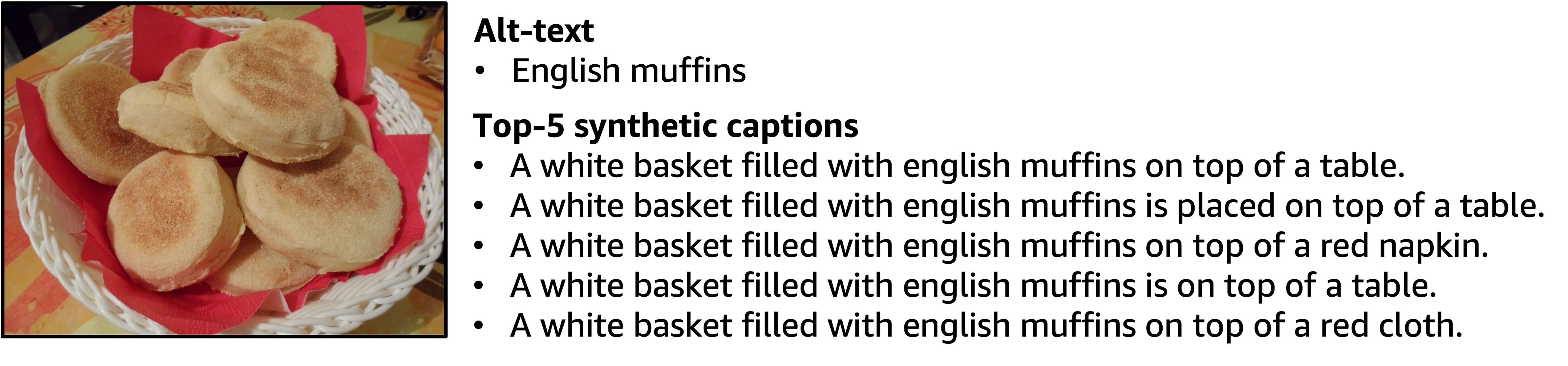}
\caption{Synthetic captions provide descriptions with more details.}
\vspace{-5mm}
\label{fig:syncap}
\end{figure}

\begin{table}[b!]
\footnotesize
\caption{Synthetic caption ablations by training SD2 for 250K steps. \textbf{IR}: ImageReward. \textbf{Top5 Syn.}: randomly select one in top-5 synthetic captions ranked by the caption prediction confidences; \textbf{Top1 Syn.}: only select top-1 synthetic caption}
\label{tab:synthetic}
\begin{tabular}{l|l|r|r|r|r}
\hline
Model & Synthetic caption & \multicolumn{1}{l|}{TIFA} & \multicolumn{1}{l|}{CLIP} & \multicolumn{1}{l|}{IR} & \multicolumn{1}{l}{FID} \\ \hline
\multirow{3}{*}{SD2} & LensArt          & 0.810                   & 0.269                   & 0.345                          & \textbf{17.9}                    \\ \cline{2-6}
                     & LensArt + Top5 Syn.              & \textbf{0.835}          & 0.270                   & \textbf{0.524}                 & 18.9           \\ \cline{2-6}
                     & LensArt + Top1 Syn.              & 0.833                   & \textbf{0.271}          & 0.513                         & 18.3                    \\ \hline
\end{tabular}
\end{table}

\subsection{Expanding Knowledge via Synthetic Captions}
To increase the valid text supervision for the smaller yet higher-quality data, we adopt an internal image captioning model, similar to BLIP2 ~\cite{blip2}, to generate synthetic captions. The captioning model produces five generic descriptions for each image ranked by prediction confidence as in Fig.~\ref{fig:syncap}. One of the 5 synthetic captions and original alt-text is randomly selected to pair with the image for model training under 50\% chance. Thus we double the image-text pairs and significantly increase the image-noun pairs as shown in Table \ref{tab:dataset}. Thanks to the text supervision expanding by synthetic captions, the image-text alignment and fidelity can be consistently boosted, as shown in Table~\ref{tab:synthetic}. 
Specifically, the ablation on LensArt shows that synthetic captions significantly improves ImageReward score. 
In addition, we find that randomly selecting one in top-5 synthetic captions is slightly better than always selecting top-1, which is adopted as the default training scheme for synthetic captions. 
Different with PixArt-$\alpha$ which always replaces the original captions with long synthetic captions, we provide an alternate way to scale up the image-text pairs by random flipping captions, which is in concurrent with the caption enhancement work of DALL-E 3 \cite{dalle3}.

\begin{figure*}[t]
\centering
\includegraphics[width=0.32\linewidth]{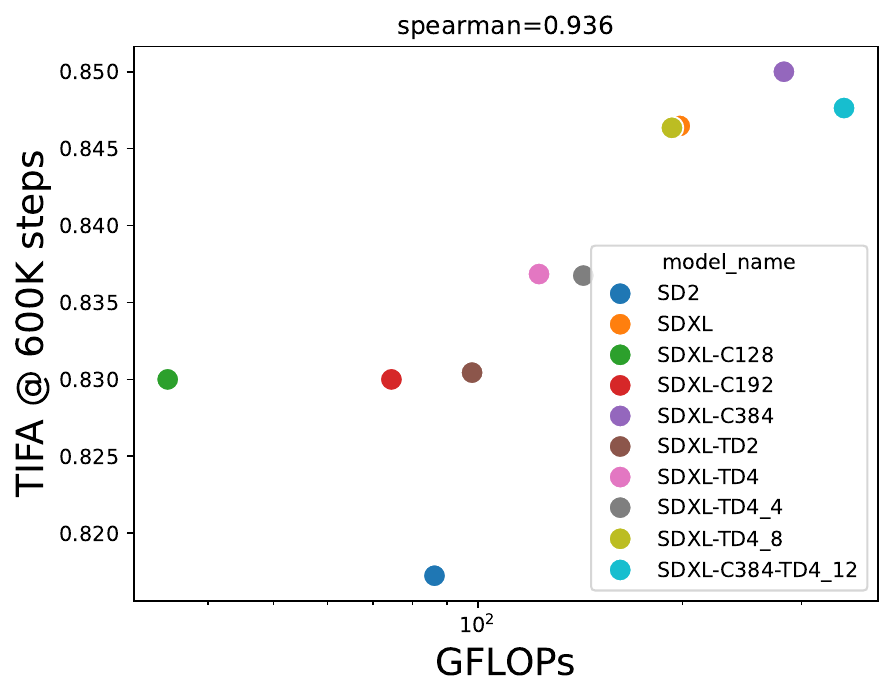}
\includegraphics[width=0.32\linewidth]{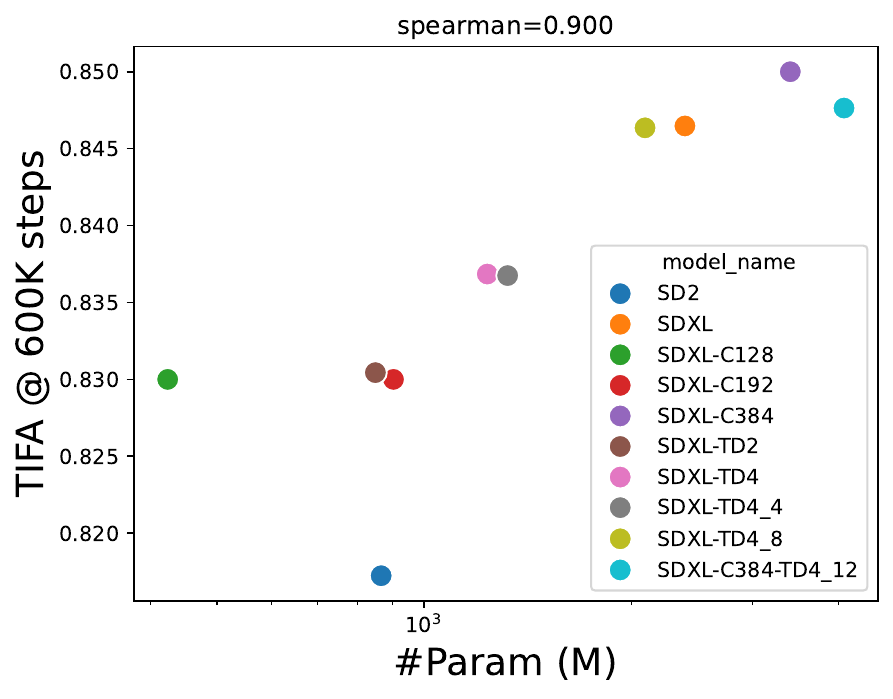}
\includegraphics[width=0.32\linewidth]{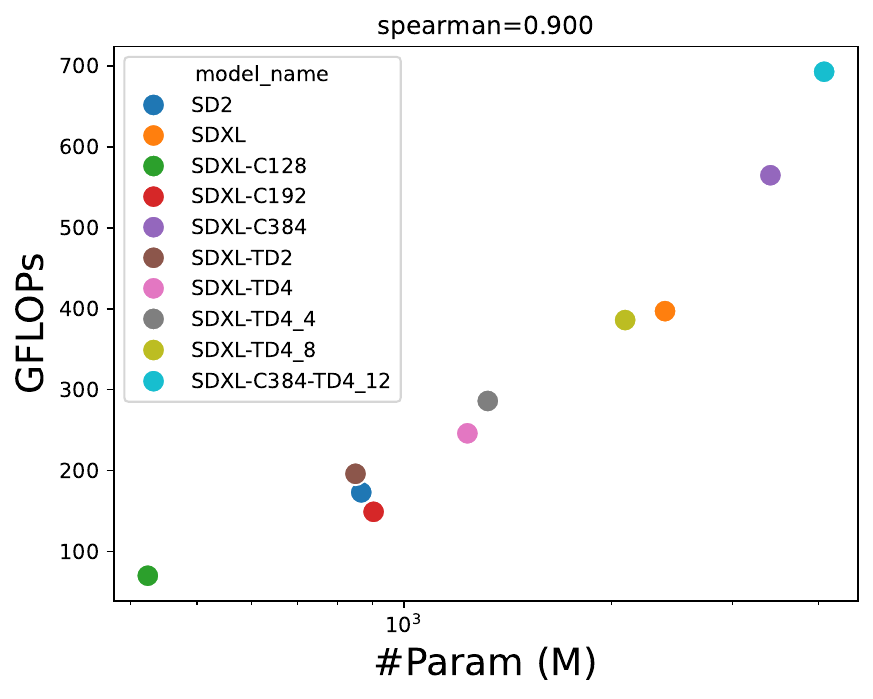}
\caption{(a-b) The correlation of TIFA score with UNet's inference complexity (GFLOPs) and number of parameters when trained with fixed steps.
(c) shows the correlation between model parameters and FLOPs.
}
\vspace{-2mm}
\label{fig:scaling_law}
\end{figure*}

\begin{figure*}[t]
\centering
\includegraphics[width=0.325\linewidth]{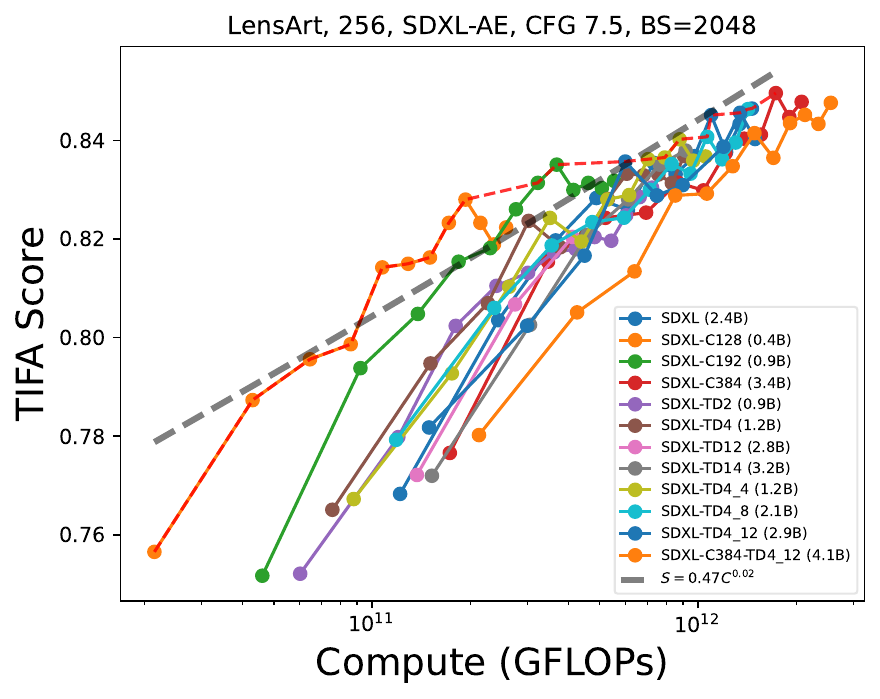}
\includegraphics[width=0.325\linewidth]{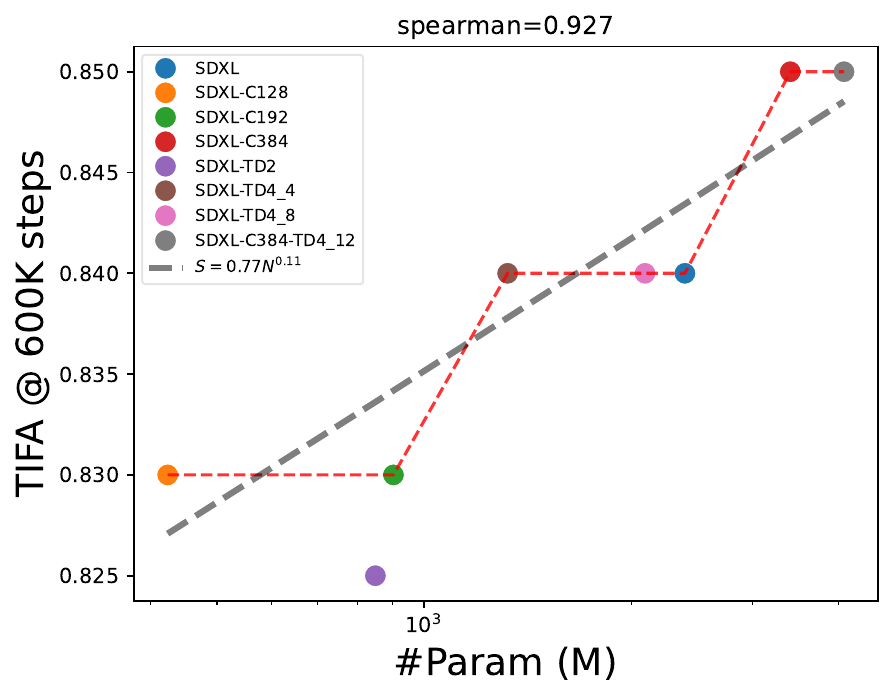}
\includegraphics[width=0.325\linewidth]{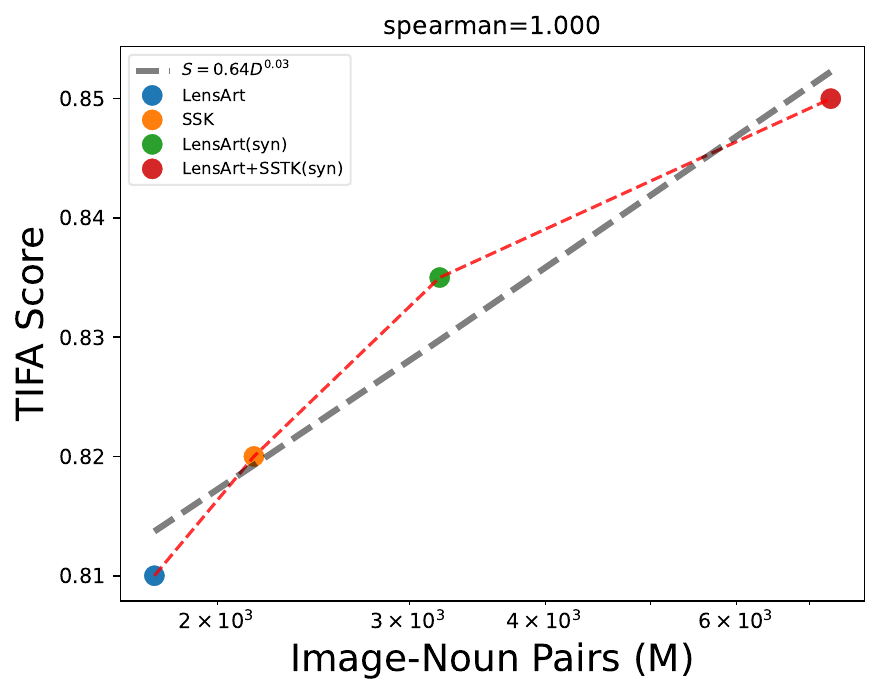}
\caption{
Fitting the scaling law of TIFA performance $S$ as a function of the training compute $C$ and model size $N$, and dataset size $D$, with the training history of SDXL variants and SD2 in fixed steps. 
The (\textcolor{red}{- -}) indicates the Pareto frontier of the scaling graph.}
\vspace{-2mm}
\label{fig:scaling_law}
\end{figure*}

\subsection{Data Scaling Increases Training Efficiency}
\paragraph{Combined datasets} The text-image alignment and image quality can be further improved as the dataset scale increases.
Here we compare the SD2 models trained on different datasets and compare their convergence speed: 1) LensArt 2) SSTK and 3) LensArt + SSTK with synthetic caption.
We also compare training with unfiltered LensArt-raw as the baseline. 
Fig.~\ref{fig:datasetcurves} shows that \emph{combining LensArt and SSTK significantly improves the convergence speed and the upper limit of two metrics in comparison with models trained on LensArt or SSTK only}.
SDXL model trained on  LensArt + SSTK reaches 0.82+ TIFA score in 100K steps, which is 2.5$\times$ faster than SDXL trained with only LensArt. 

\paragraph{Advanced models scale better on larger dataset}
Fig.~\ref{fig:scale_model_dataset} shows that SD2 model can get significant performance gain when training on the scaled (combined) dataset. 
SDXL still gets performance gain over the SD2 model when trained with the scaled dataset, indicating that models with large capacity have better performance when the dataset scale increases.

%% file: sec/6_scaling_law.tex
\section{More Scaling Properties}
\paragraph{Relationship between performance and model FLOPs}
Fig.~\ref{fig:scaling_law} (a-b) shows the correlation between TIFA score obtained at fixed steps (i.e., 600K) and model compute complexity (GFLOPs) as well as model size (\#Params) for all examined SD2 and SDXL variants.
We see the TIFA score correlates slightly better with FLOPs than parameters, indicating the importance of model compute when training budget is sufficient, which aligns with our findings in Sec~\ref{sec:unet}.

\paragraph{Relationship between performance and data size}
Fig.~\ref{fig:scaling_law}(c) shows the correlation between SD2's TIFA score and dataset size in terms of number of image-noun pairs. Each image-noun pair is defined as an image paired with one noun in its caption. It measures the interaction between the fine-grained text unit with the image. 
We see a linear correlation between TIFA and the scale of image-noun pairs when scaling up the cleaned data. 
Compared to LensArt-raw with similar amount of image-noun pairs, LensArt+SSTK is much better, which indicates the importance of data quality.

\paragraph{Numerical Scaling Law}
The scaling law of LLMs~\cite{kaplan2020scaling, hoffmann2022training} reveals that LLM's performance has a precise power-law scaling as a function of dataset size, model size, and compute budget. 
Here we fit similar scaling functions for SDXL variants and SD2.
The TIFA score $S$ can be a function of total compute $C$ (GFLOPs), model parameter size $N$ (M parameters) and dataset size $D$ (M image-noun pairs) as shown in Fig.~\ref{fig:scaling_law}. 
Specifically, with the Pareto frontier data points, we can fit the power-law functions to be
$S=0.47C^{0.02}$, $S=0.77N^{0.11}$, and 
$S=0.64D^{0.03}$, which approximate the performance in a range given sufficient training.
Similar as LLMs, we see larger models are more sample efficient and smaller models are more compute efficient.

\paragraph{Model Evaluation at Low Resolution}
One may wonder whether the models' relative performance will change at high resolution training, so that the gap between models can be mitigated. 
In the Appendix, we show continue training models at 512 resolution slightly improve their 256 resolution metrics without significant change. 
Though image quality and aesthetics can be improved via high-quality fine-tuning~\cite{emu}, it is hard for the inferior model to surpass when trained on the same data, especially when the high resolution data is much less than its lower resolution version. 
The majority composition capability is developed at low resolution, which enables us to assess model's performance at the early stage of low resolution training.

%% file: sec/7_conclusion.tex
\section{Conclusions}
We present a systematic study on the scaling properties of training diffusion based T2I models, including the effects of scaling both denoising backbone and dataset.
Our study demonstrates practical paths to improve T2I model performance by properly scaling up existing denoising backbones with large-scale datasets, which results in better text-image alignment and image quality, as well as training efficiency. 
We hope those findings can benefit the community for pursuing more scaling-efficient models.

%% file: sec/X_suppl.tex
\onecolumn
{
\centering
\Large
\textbf{\thetitle}\\
\vspace{0.5em}Supplementary Material \\
\vspace{1.0em}
}

\section{Evaluation Details}
\label{sec:app_eval}
\paragraph{Prompts} We generate images with two prompt sets for evaluation: 1) 4081 prompts from TIFA~\cite{tifa} benchmark. The benchmark contains questions about 4,550 distinct elements in 12 categories, including \textit{object}, \textit{animal/human}, \textit{attribute}, \textit{activity}, \textit{spatial}, \textit{location}, \textit{color},
\textit{counting}, \textit{food}, \textit{material}, \textit{shape}, and
\textit{other}. 2) randomly sampled 10K prompts from  MSCOCO~\cite{mscoco} 2014 validation set.

\paragraph{Metrics}
In addition to previously introduced  TIFA~\cite{tifa} and ImageReward~\cite{imagereward} scores, we also calculate the following metrics:
\begin{itemize}
    \item \textbf{FID}: FID measures the fidelity or similarity of the generated images to the groundtruth images. The score is calculated based on the MSCOCO-10K prompts and their corresponding images. We resize the groundtruth images to the same resolution (256$\times$256 or 512$\times$512) as the generated images.
    \item \textbf{CLIP}: The CLIP score~\cite{clip, clipscore} measures how the generated image aligns with the prompt. Specifically, the cosine similarity between the CLIP embeddings of the prompt and the generated image. Here we calculate it with the MSCOCO-10K prompts and report the average value. 
    \item \textbf{Human Preference Score (HPS)}~\cite{hpsv2}: HPSv2 is a preference prediction model trained with human preference. We calculate the scores based on the TIFA prompts and report the average value.
\end{itemize}

\paragraph{Inference Settings} Given a prompt set and a pre-trained model, we generate images at 256$\times$256 resolution with DDIM~\cite{song2020denoising} 50 steps, using default CFG 7.5 and fixed seed for all prompts.
For each model checkpoint, we use its non-EMA weights for evaluation.

\section{More Results on UNet Scaling} 
\begin{figure*}[htb]
\centering
\includegraphics[width=0.195\linewidth]{figures/sd2_256_sdxl_ae_unet_lensartv1_channels_tifa_steps.pdf}
\includegraphics[width=0.195\linewidth]{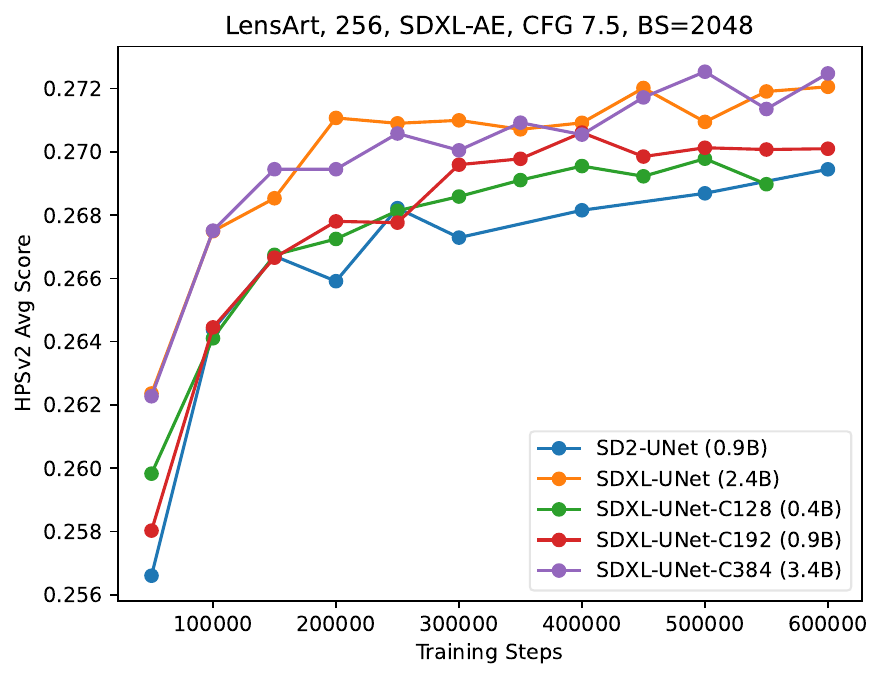}
\includegraphics[width=0.195\linewidth]{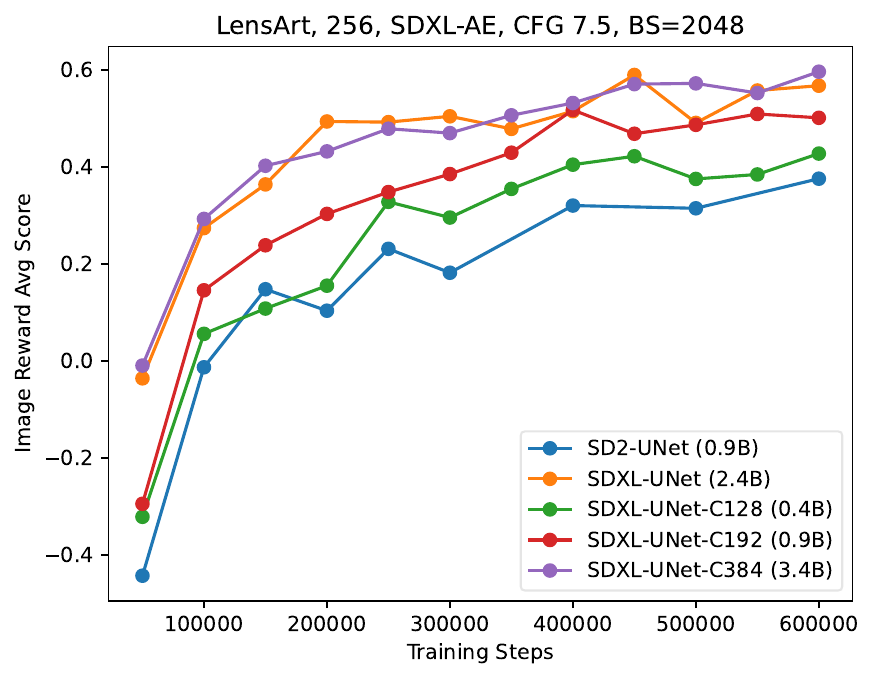}
\includegraphics[width=0.195\linewidth]{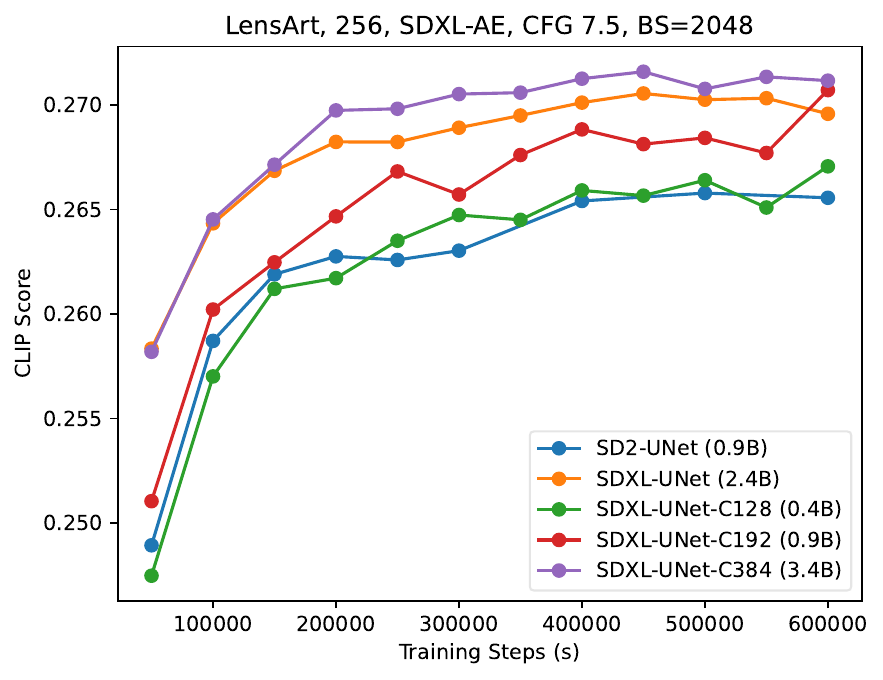}
\includegraphics[width=0.195\linewidth]{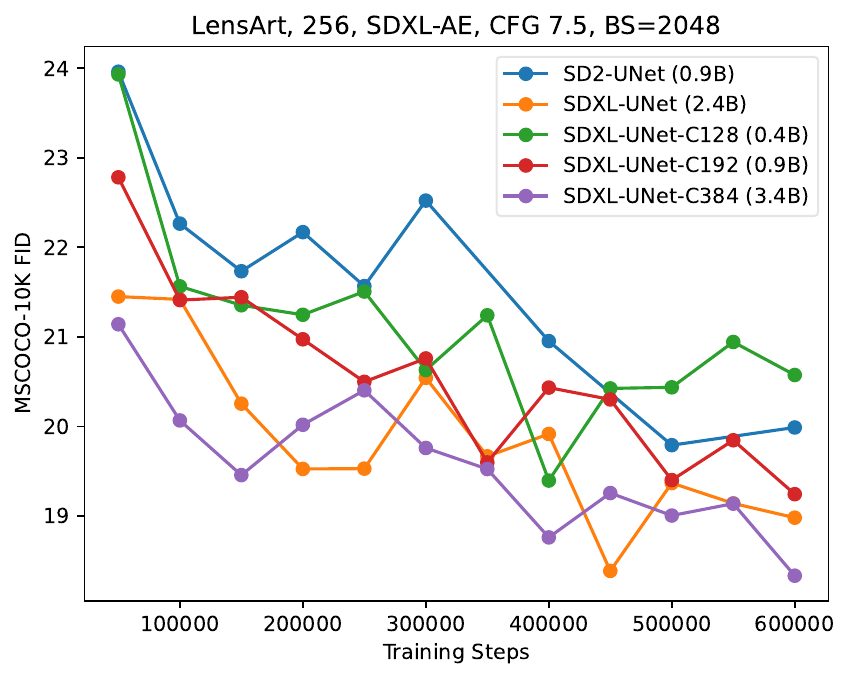}\\
\includegraphics[width=0.195\linewidth]{figures/sd2_256_sdxl_ae_unet_lensartv1_td_tifa_steps.pdf}
\includegraphics[width=0.195\linewidth]{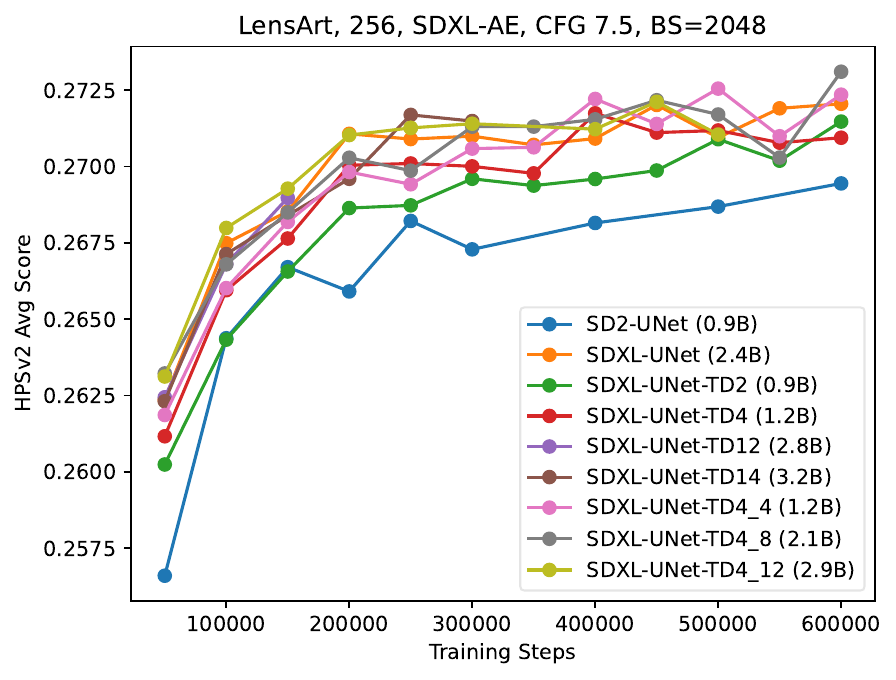}
\includegraphics[width=0.195\linewidth]{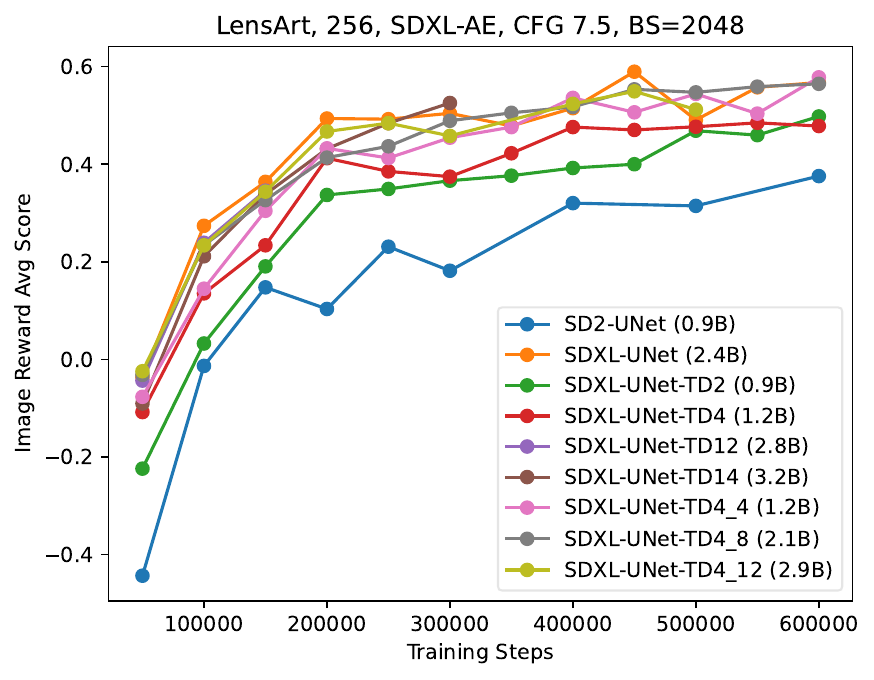}
\includegraphics[width=0.195\linewidth]{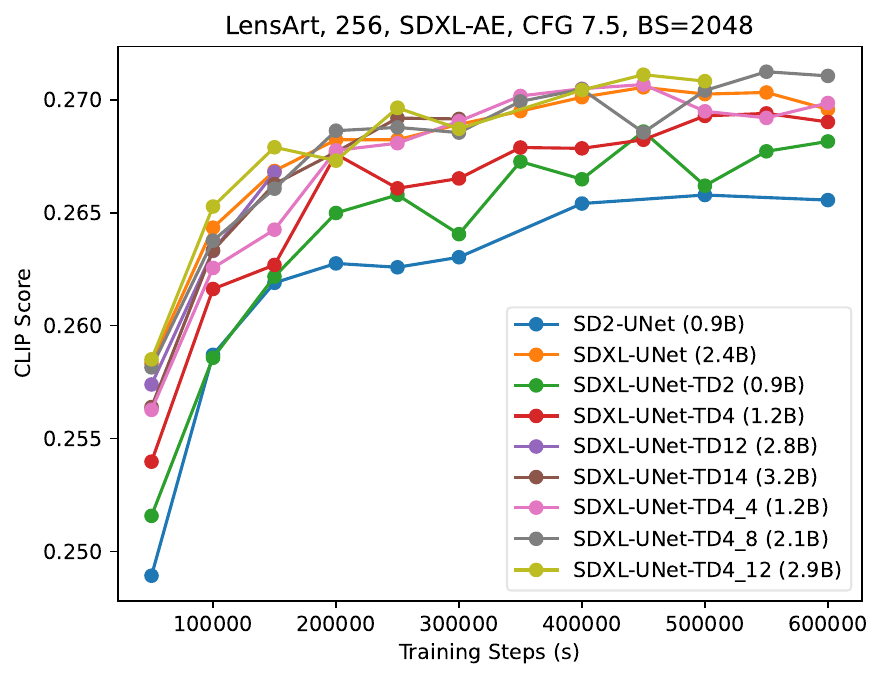}
\includegraphics[width=0.195\linewidth]{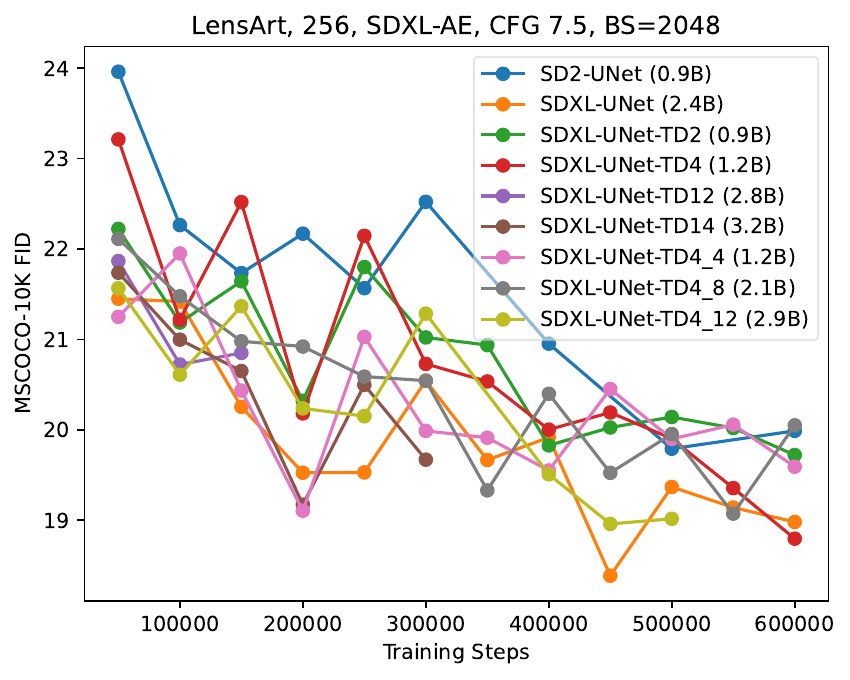}\\
\includegraphics[width=0.195\linewidth]{figures/sd2_256_sdxl_ae_unet_lensartv1_td_channels_tifa_steps.pdf}
\includegraphics[width=0.195\linewidth]{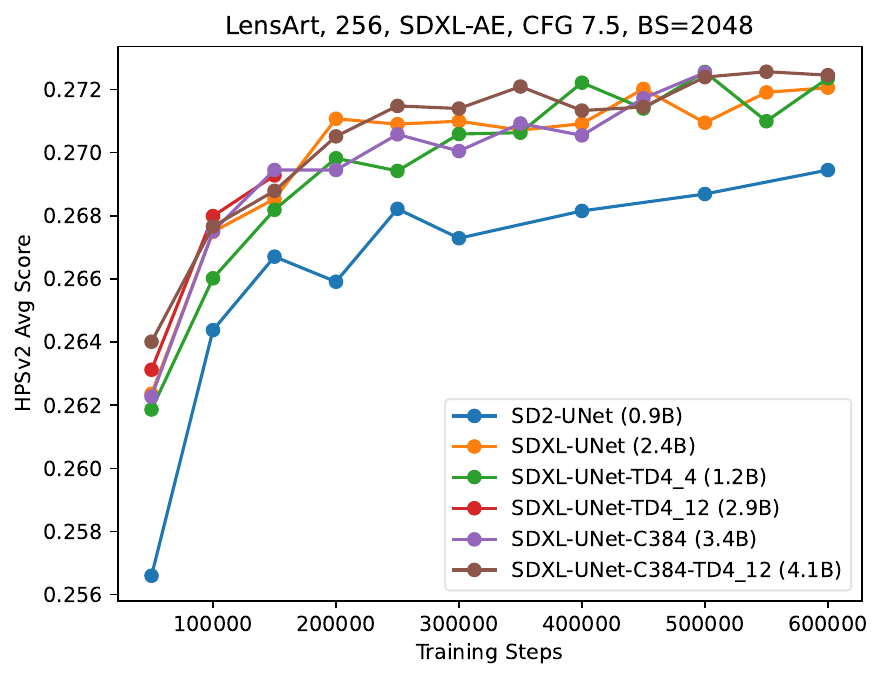}
\includegraphics[width=0.195\linewidth]{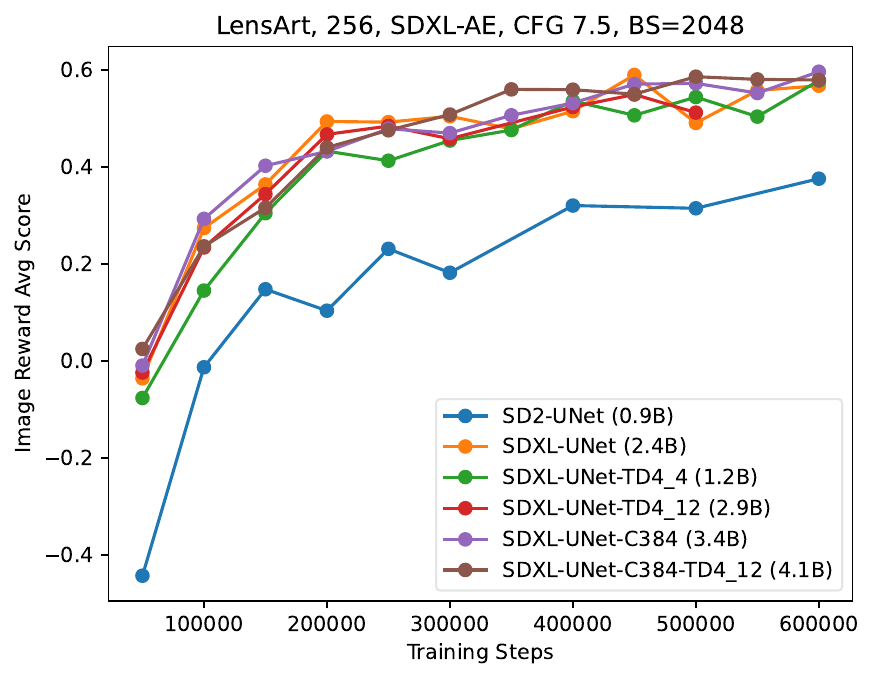}
\includegraphics[width=0.195\linewidth]{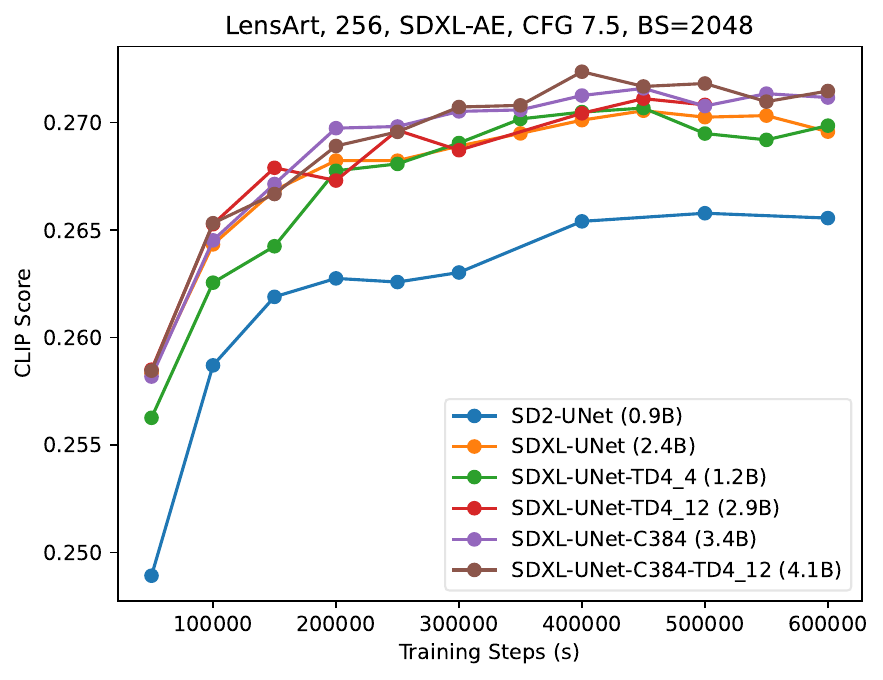}
\includegraphics[width=0.195\linewidth]{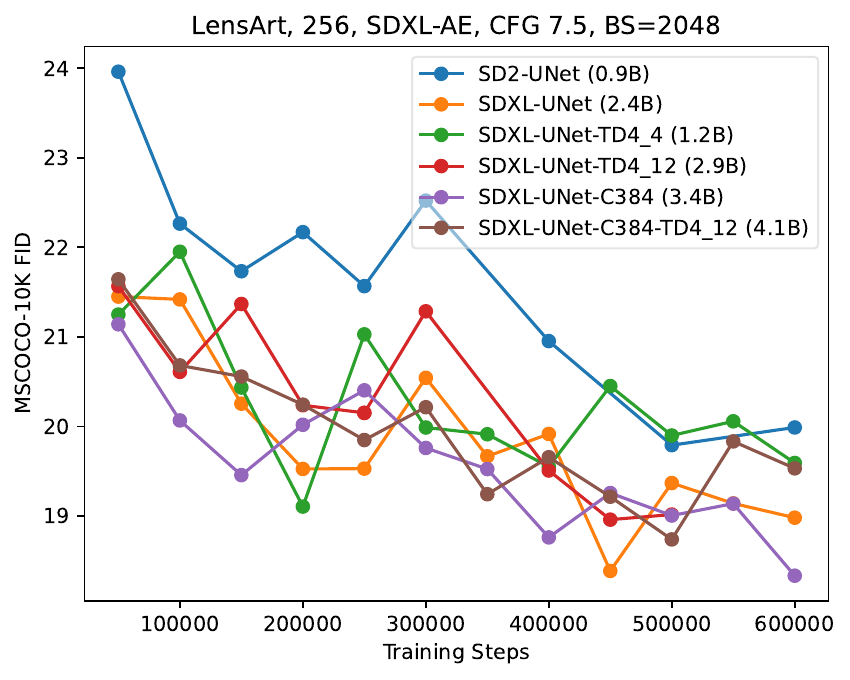}\\

\caption{The evolution of all metrics during training for UNet variants. 
The baseline models are the UNets of SD2 and SDXL. 
All models are trained with SDXL VAE at 256$\times$256 resolution.
The 1st row shows SDXL UNets with different initial channels.
The 2nd row shows SDXL UNets with different TDs.
The 3rd row compares SDXL UNets with both increased channels and TDs.
}
\label{fig:more_metrics}
\end{figure*}

\paragraph{Evolution of all metrics for UNet variants} We have shown the TIFA evolution curves of SDXL~\cite{sdxl} UNet variants in Sec.~3.
Here we show the evolution of other metrics during training for all UNet variants in Fig.~\ref{fig:more_metrics}, including the change of \textit{channels}, \textit{transformer depth} and both of them.
The pattern of other metrics is very similar as TIFA and the relative performance among models is stable across metrics, e.g., the 1st row of Fig.~\ref{fig:more_metrics} shows that UNets with more channels tend to have better TIFA, HPSv2,  ImageReward, CLIP, and FID scores. 
Though FID score has more variations during training.

\begin{figure}[h!]
\centering
\includegraphics[width=0.195\linewidth]{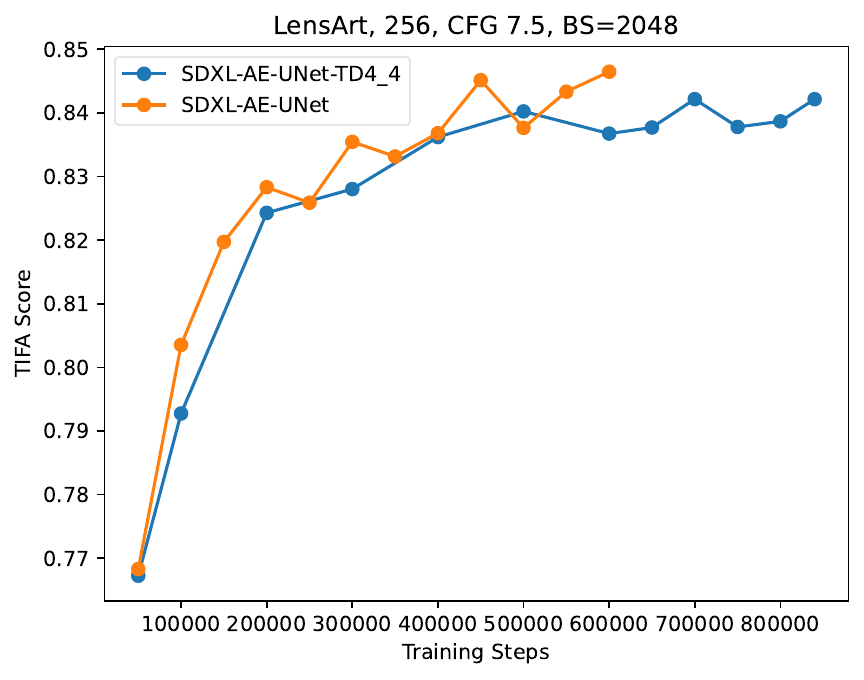}
\includegraphics[width=0.195\linewidth]{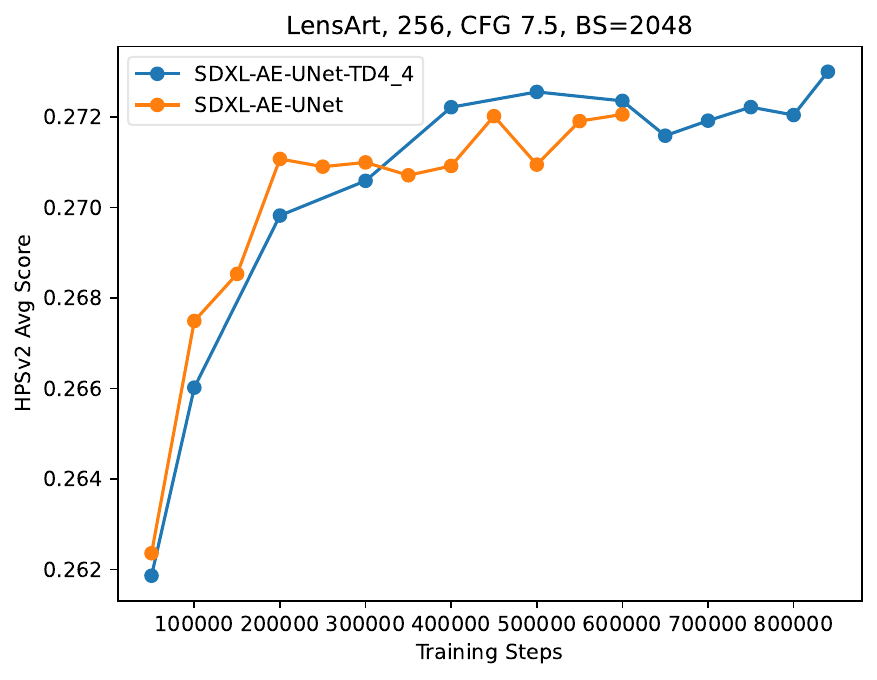}
\includegraphics[width=0.195\linewidth]{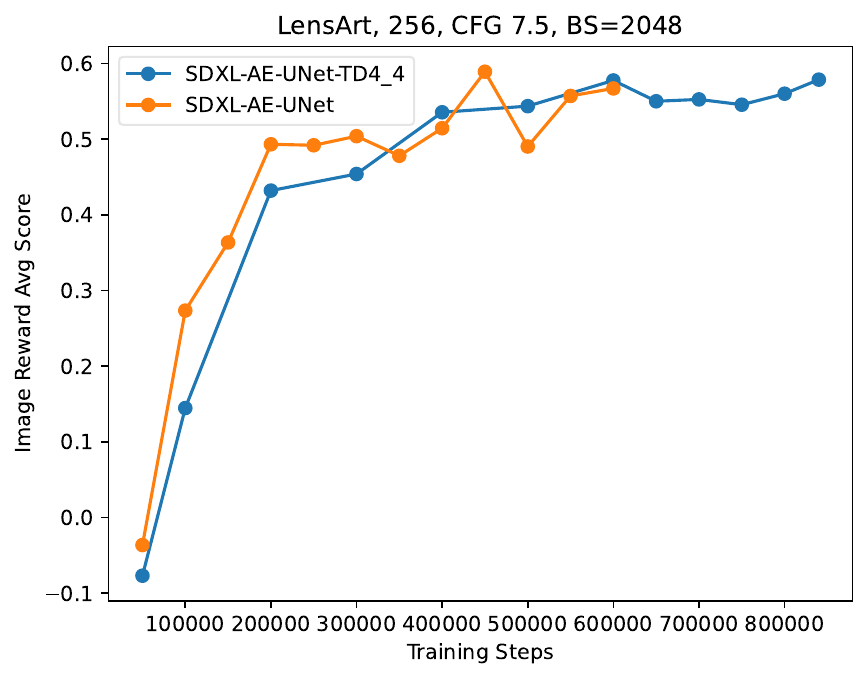}
\includegraphics[width=0.195\linewidth]{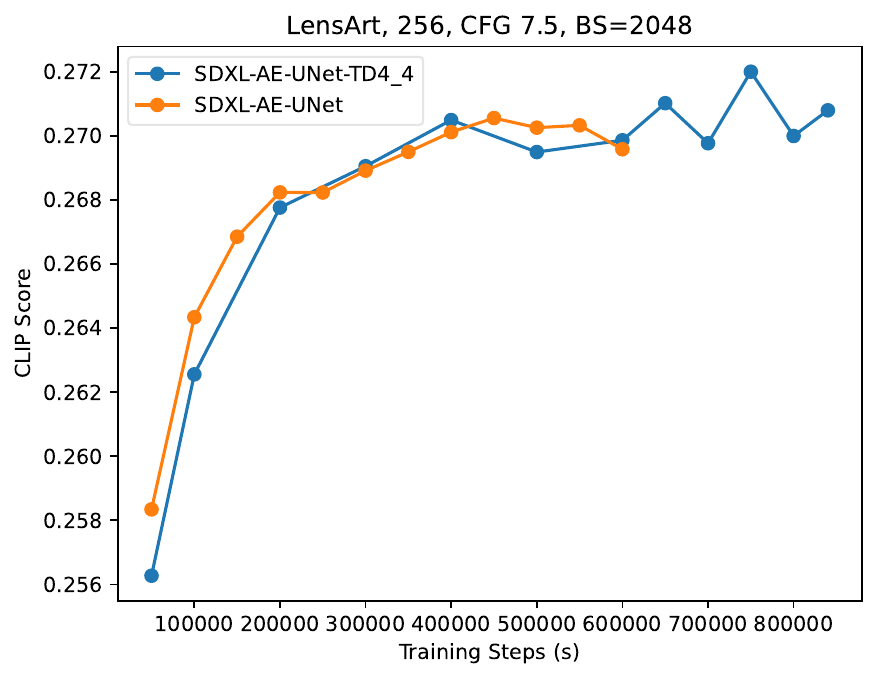}
\includegraphics[width=0.195\linewidth]{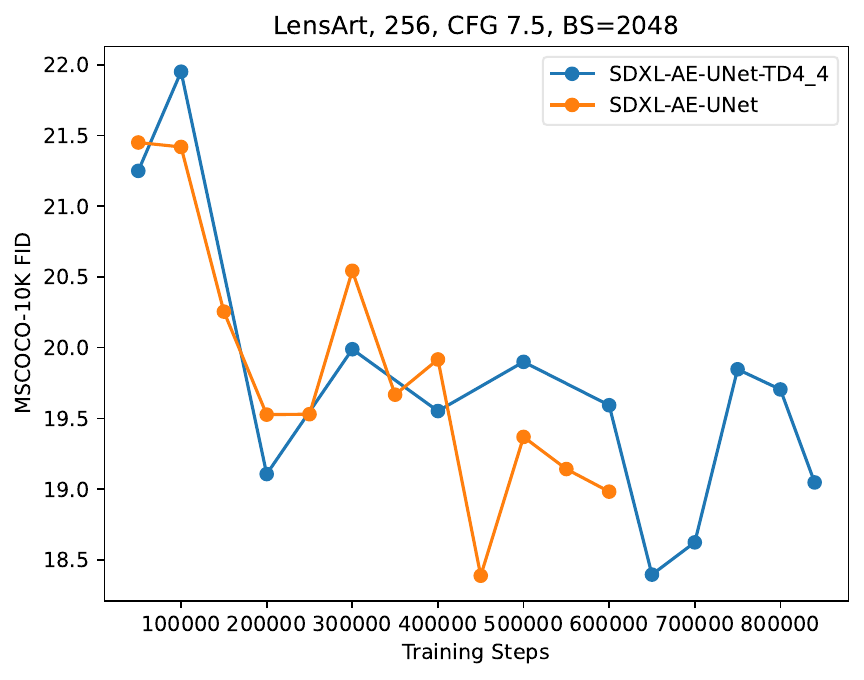}
\\
\includegraphics[width=0.195\linewidth]{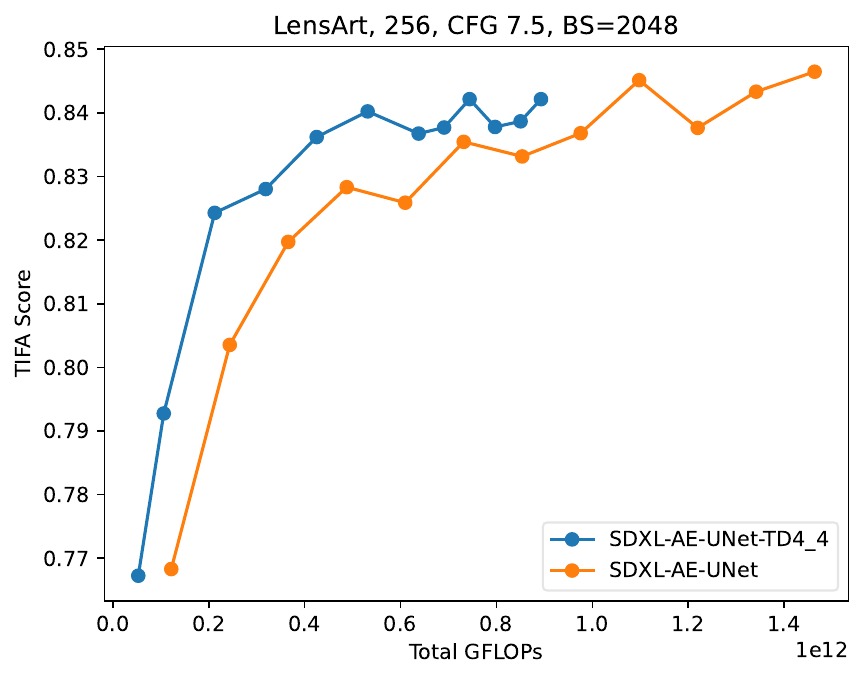}
\includegraphics[width=0.195\linewidth]{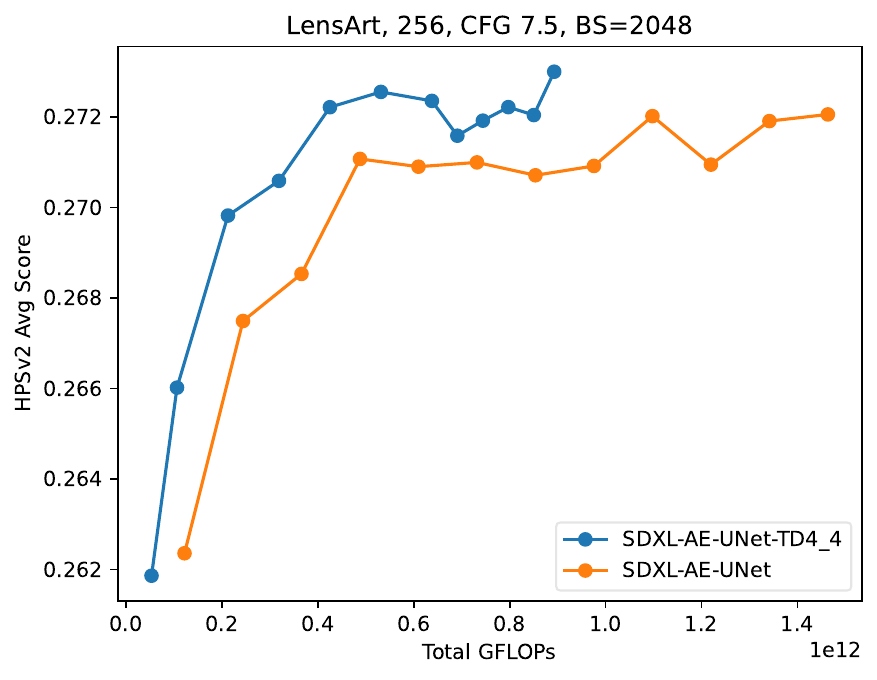}
\includegraphics[width=0.195\linewidth]{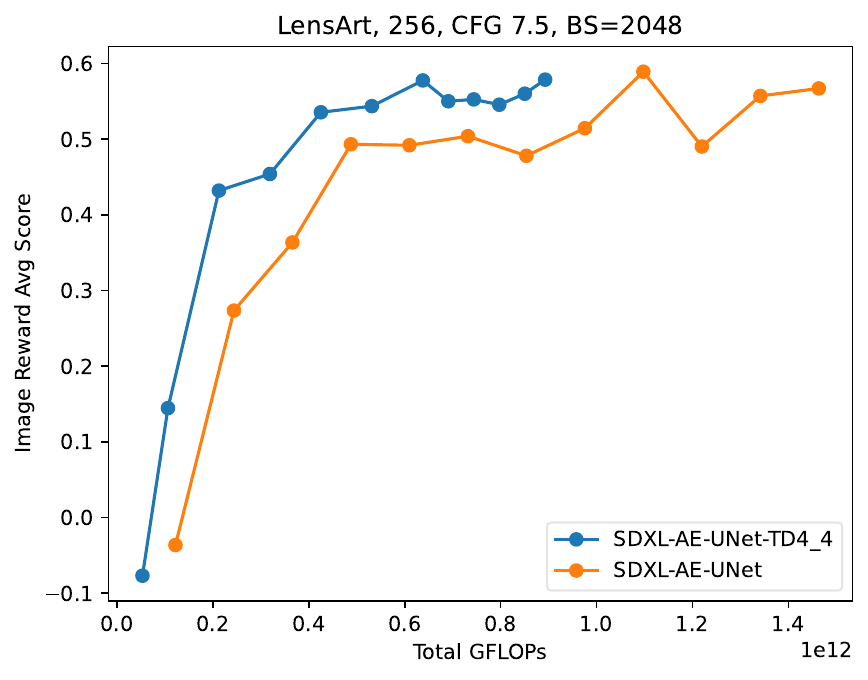}
\includegraphics[width=0.195\linewidth]{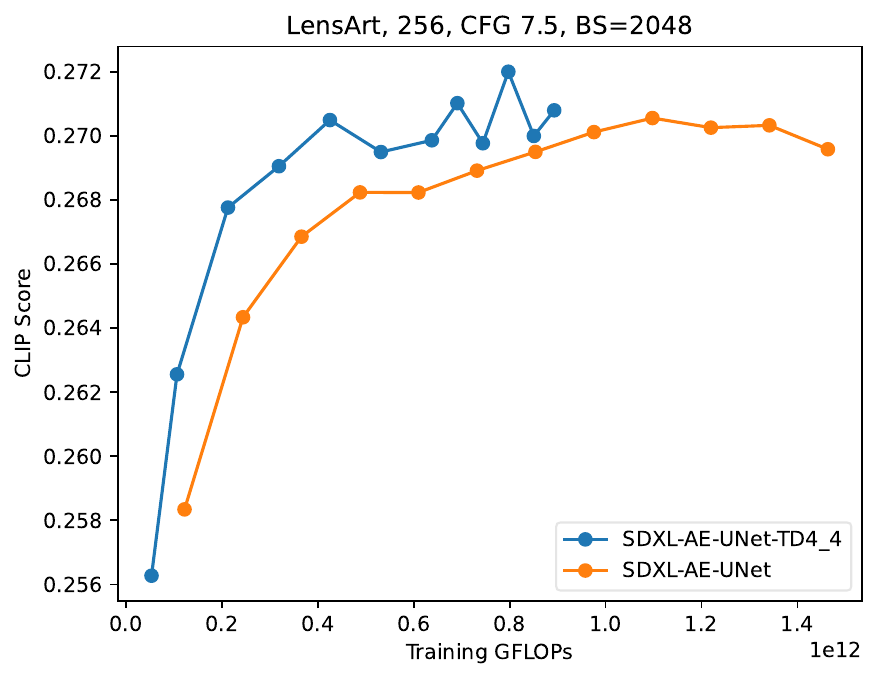}
\includegraphics[width=0.195\linewidth]{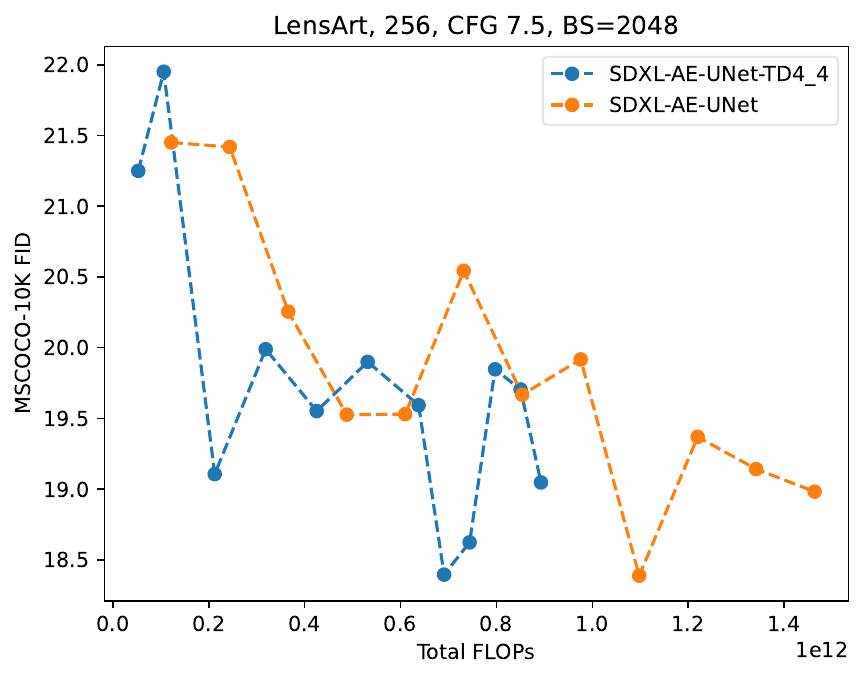}
\caption{
\small Comparing metrics evolution speed of SDXL UNet and its TD4\_4 variant in terms of training steps and total compute (GFLOPs).
TD4\_4 achieves similar or better metric scores at much less training cost.}
\label{fig:td4_4_longer}
\end{figure}

\paragraph{Comparing the training efficiency of SDXL UNet and its variant}
Previously we introduce a smaller SDXL UNet variant, i.e., TD4\_4, which is 45\% smaller, 28\% faster, and has competitive performance as SDXL-UNet when trained with the same steps (Fig.~\ref{fig:more_metrics}).
Here we compare their metrics in terms of training steps as well as the total compute (GFLOPs).
We extend the training steps of TD4\_4 from 600K to 850K to see whether the performance can be further improved.
As shown in Fig.~\ref{fig:td4_4_longer}, TD4\_4 achieves similar or better metrics in comparison with SDXL UNet with much less computation cost.
It suggests that TD4\_4 is a more compute efficient model when the training budget is limited.

\vspace{-2mm}

\section{More Results on Dataset Scaling} 
\begin{figure*}[h!]
\centering
\includegraphics[width=0.195\linewidth]{figures/sd2_256_datasets_public_tifa_steps.pdf}
\includegraphics[width=0.195\linewidth]{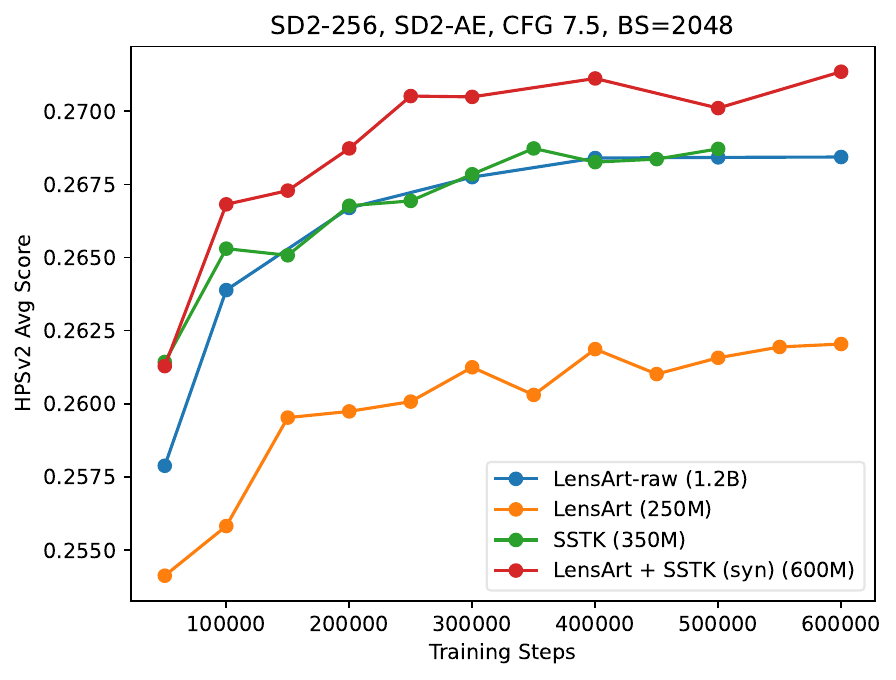}
\includegraphics[width=0.195\linewidth]{figures/sd2_256_datasets_public_image_reward_avg_steps.pdf}
\includegraphics[width=0.195\linewidth]{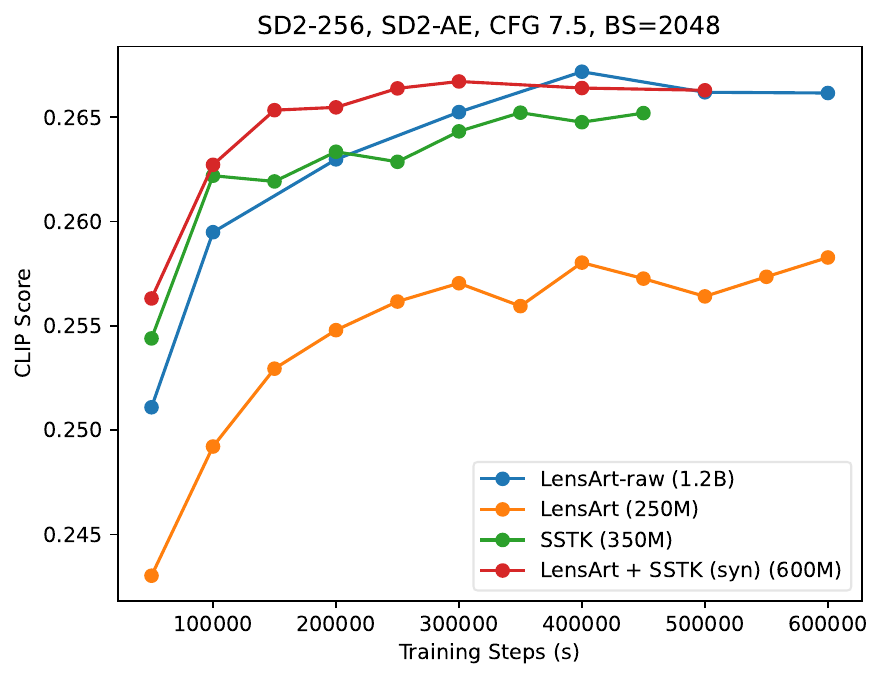}
\includegraphics[width=0.195\linewidth]{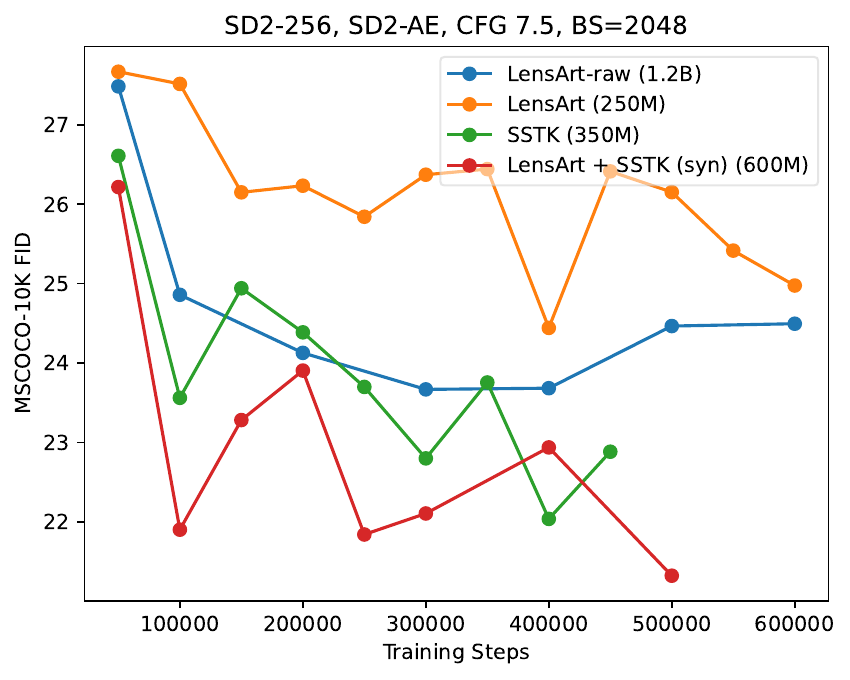}
\caption{Training SD2 model with different datasets. All metrics show that LensArt + SSTK has better scores than LensArt or SSTK only. Note that the HPSv2 and CLIP scores for LensArt-Raw are much higher than LensArt. 
The reason is that unfiltered dataset tends to generate images with more meme text.
}
\label{fig:dataset_curves_more}
\end{figure*}

\paragraph{Evolution of all metrics for SD2-UNet trained on different datasets}
We have shown the TIFA and ImageReward evolution curves of SD2-UNet trained on different datasets in Sec.~4.
Here we show the evolution of all metrics in Fig.~\ref{fig:dataset_curves_more}.
The trend of other metrics is similar as TIFA, except the HPSv2 and CLIP scores for \textit{LensArt-Raw}, which have higher values than \textit{LensArt}. 
We find the reason is that the \textit{LensArt-Raw} model tend to generate images with more meme text due to a large amount of data has such patterns, and such images usually results in higher values on those two metrics.
Those metrics become more precise and meaningful after the training data is filtered by removing those meme images.

\section{The Effect of VAE Improvement}
\label{sec:vae}
\begin{figure*}[h!]
\centering
\includegraphics[width=0.195\linewidth]{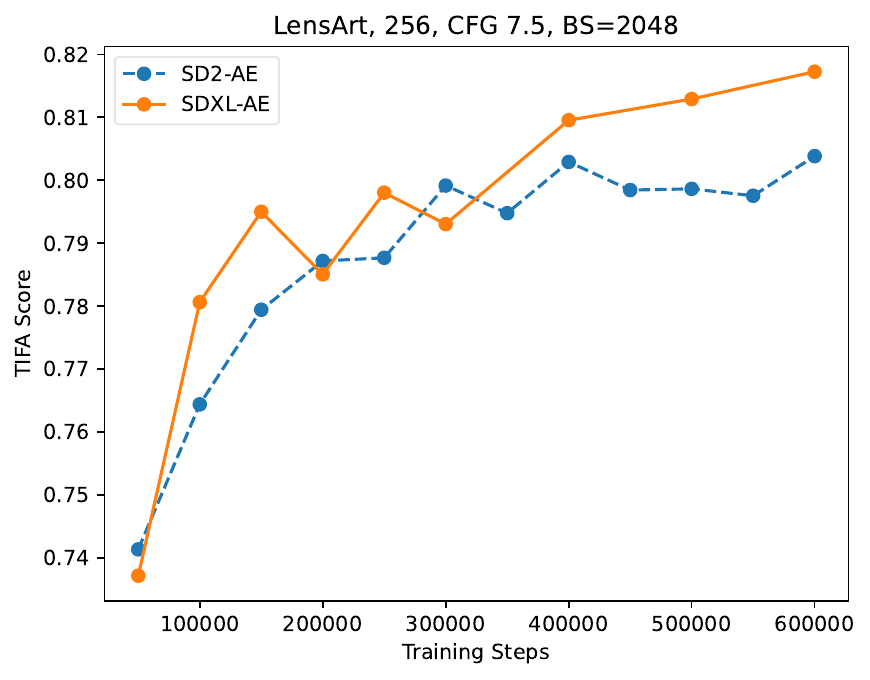}
\includegraphics[width=0.195\linewidth]{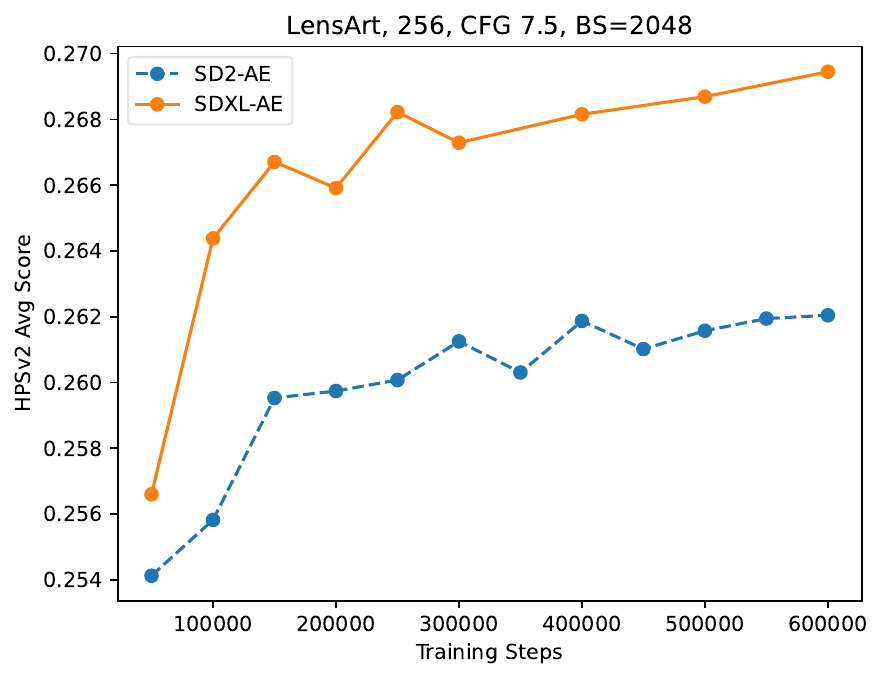}
\includegraphics[width=0.195\linewidth]{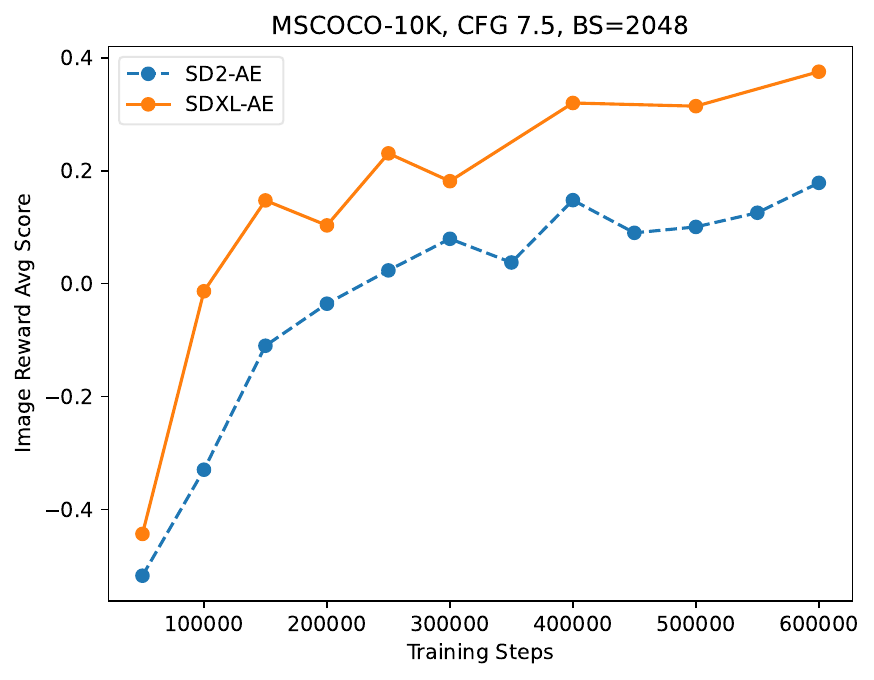}
\includegraphics[width=0.195\linewidth]{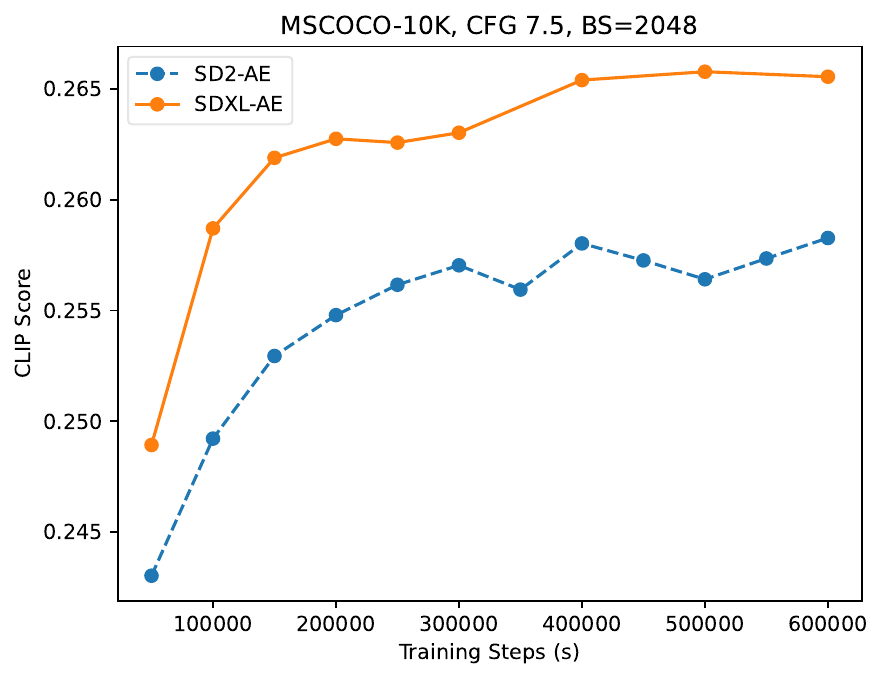}
\includegraphics[width=0.195\linewidth]{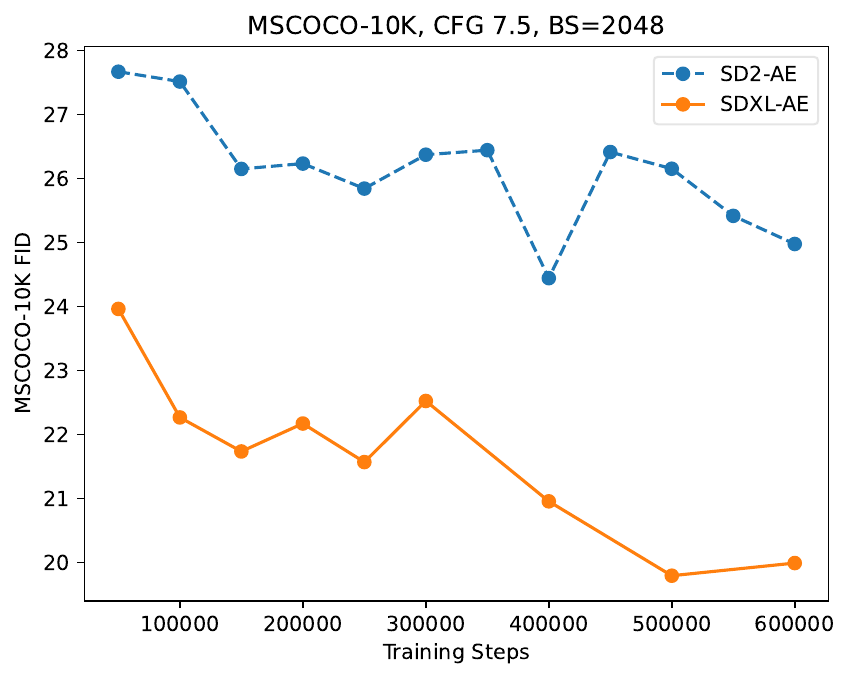}
\caption{Training SD2 UNet model with different VAEs. The SDXL's VAE has significant improvement on all metrics over SD2's VAE.}
\label{fig:vae}
\end{figure*}

SDXL~\cite{sdxl} introduced a better trained VAE and shows improved reconstruction metrics in comparison with its SD2 version.
However, the impacts on the evaluation metrics are not fully explored.
Here we ablate the effect of VAE on the evaluation metrics.
We compare the training of same SD2-UNet with different VAEs, i.e., SD2's VAE and SDXL's VAE, while keeping all other settings the same.
Fig.~\ref{fig:vae} shows that the improvement of SDXL's VAE over SD2's VAE is significant for all metrics.

\section{Scaling the Batch Size}
\begin{figure*}[h!]
\centering
\includegraphics[width=0.195\linewidth]{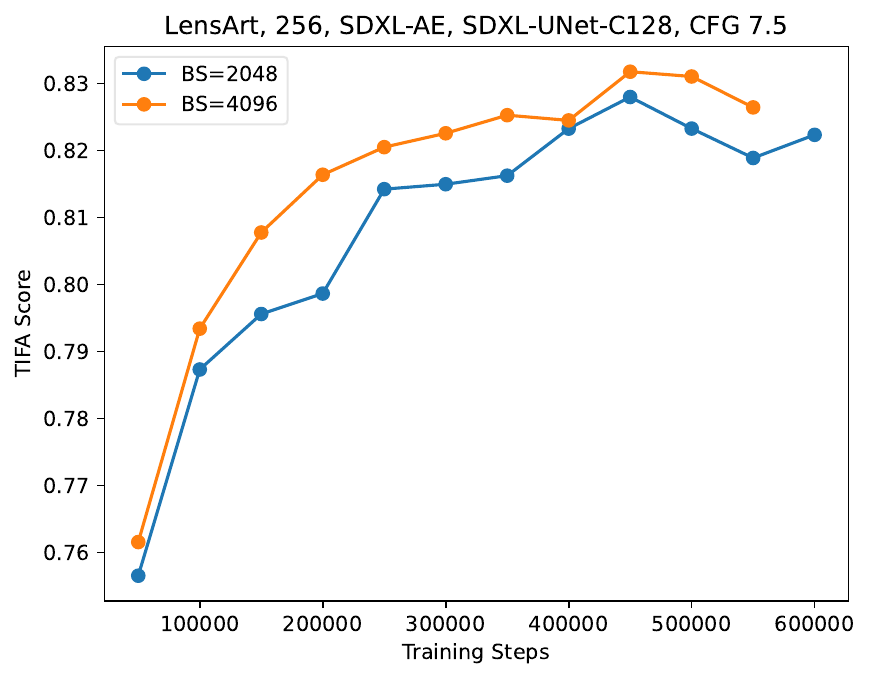}
\includegraphics[width=0.195\linewidth]{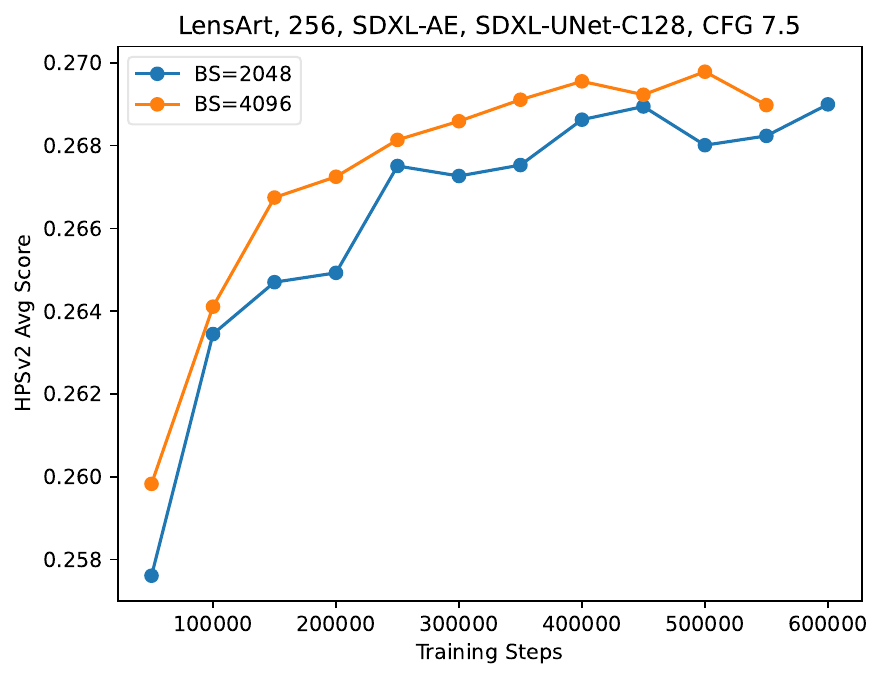}
\includegraphics[width=0.195\linewidth]{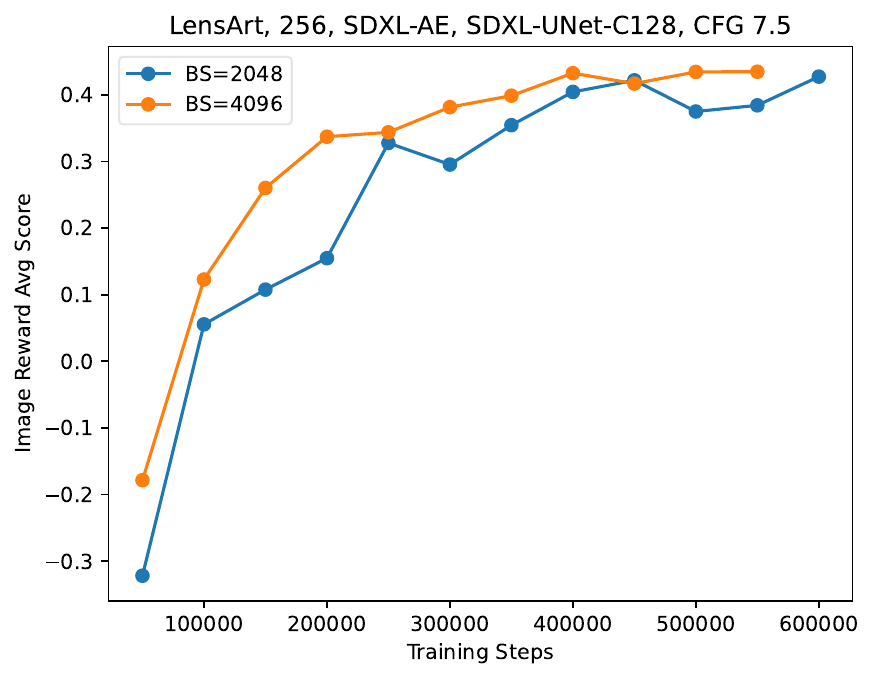}
\includegraphics[width=0.195\linewidth]{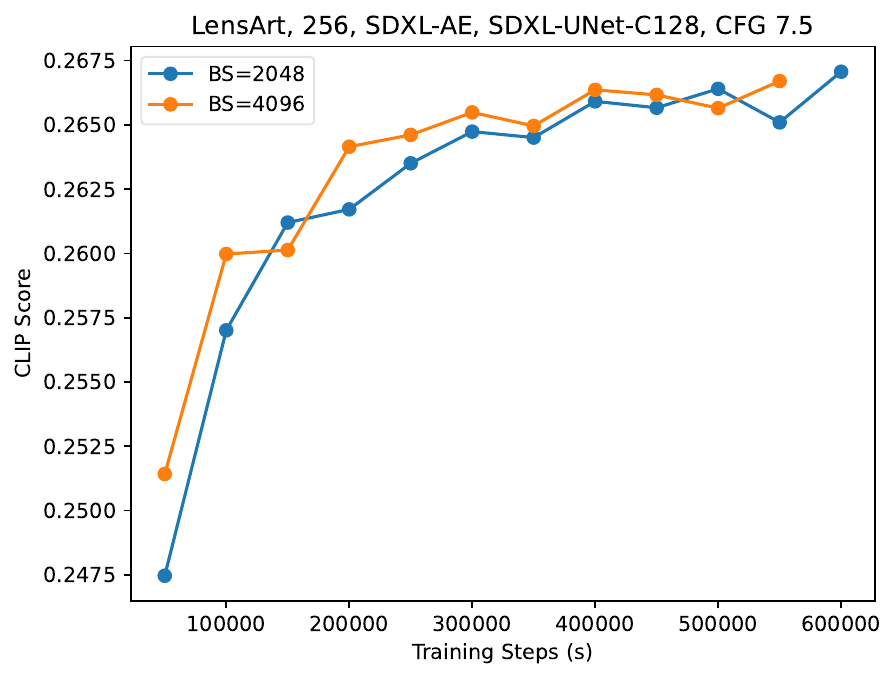}
\includegraphics[width=0.195\linewidth]{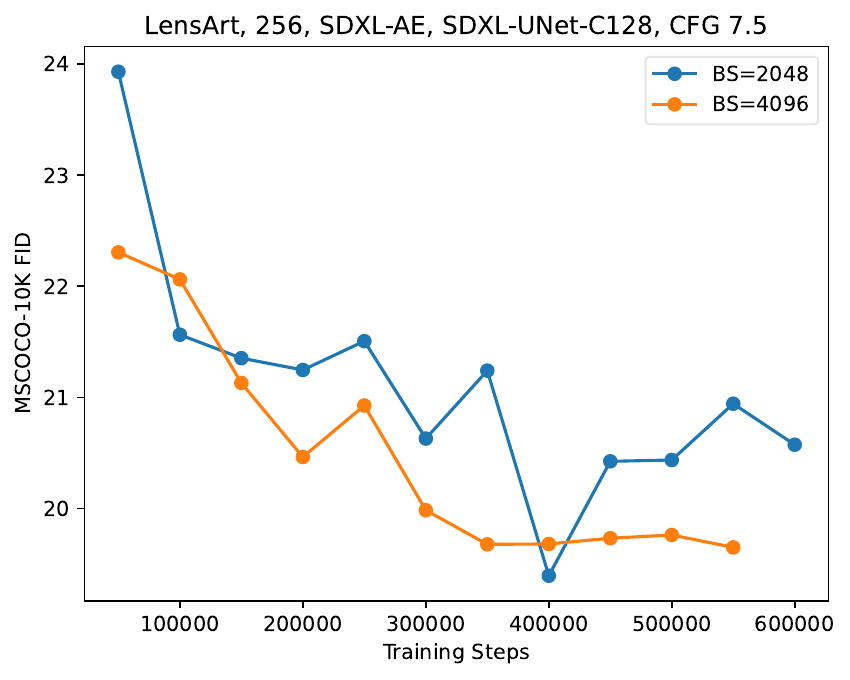}
\caption{Training SDXL-UNet-C128 with different batch sizes.}
\label{fig:batch_size}
\end{figure*}

To scale out the training of large diffusion models with more machines, increasing batch size is usually an effective approach. 
We have been using consistent batch size 2048 in all experiments for controlled studies.
Here we also show the effect of batch size on the evolution of metrics.
We compare the training of SDXL UNet with 128 channels in different batch sizes, i.e., 2048 and 4096, while keeping other training configs the same.
Fig.~\ref{fig:batch_size} shows that larger batch size yields better metrics in terms of same iteration numbers.
The convergence curve of FID score is more smooth than smaller batch size.

\section{Model Evaluation at Low Resolution Training}

\begin{figure}[h!]
\centering
\includegraphics[width=0.32\linewidth]{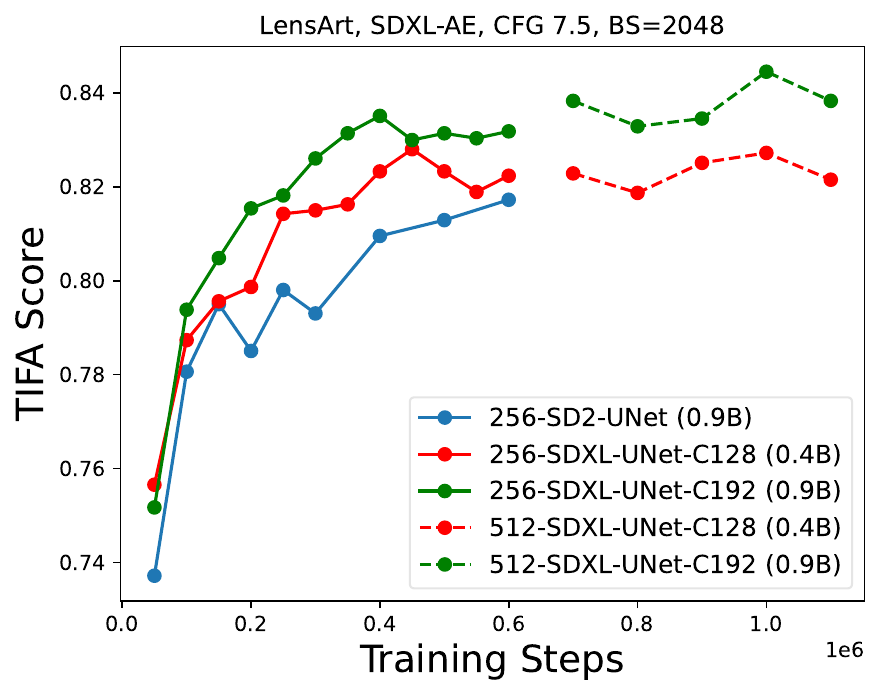}
\includegraphics[width=0.32\linewidth]{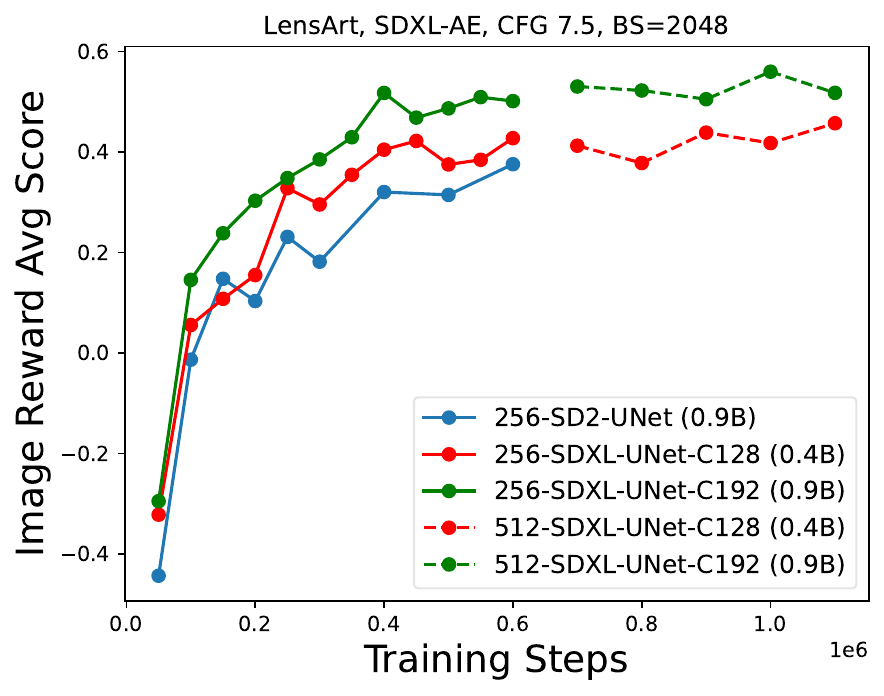}
\caption{TIFA and ImageReward do not change much during high resolution fine-tuning stage (dashed lines)
}
\label{fig:tifa_res}
\end{figure}

The evaluation metrics at 256 resolution can provide early signals on their performance at high resolutions, which is informative for quick model ablation and selection.
The reason is that the high resolution training usually utilizes a subset of images of the dataset, and the text-image alignment and image quality scores usually do not change significantly once they are fully trained at lower resolution, especially the text-image alignment performance.
Given two well trained SDXL models (C128 and C192) at 256 resolution, which has clear performance gap, we continue training them at 512 resolution and measure their performance gap. 
As shown in Fig.~\ref{fig:tifa_res},  both two SDXL UNet models can get performance improvement at 512 resolution, but C128 model still yields worse performance than C192.

\section{Caption Analysis}
\label{sec:dataset_analysis}
\begin{figure*}[h]
\centering
\includegraphics[width=0.4\linewidth]{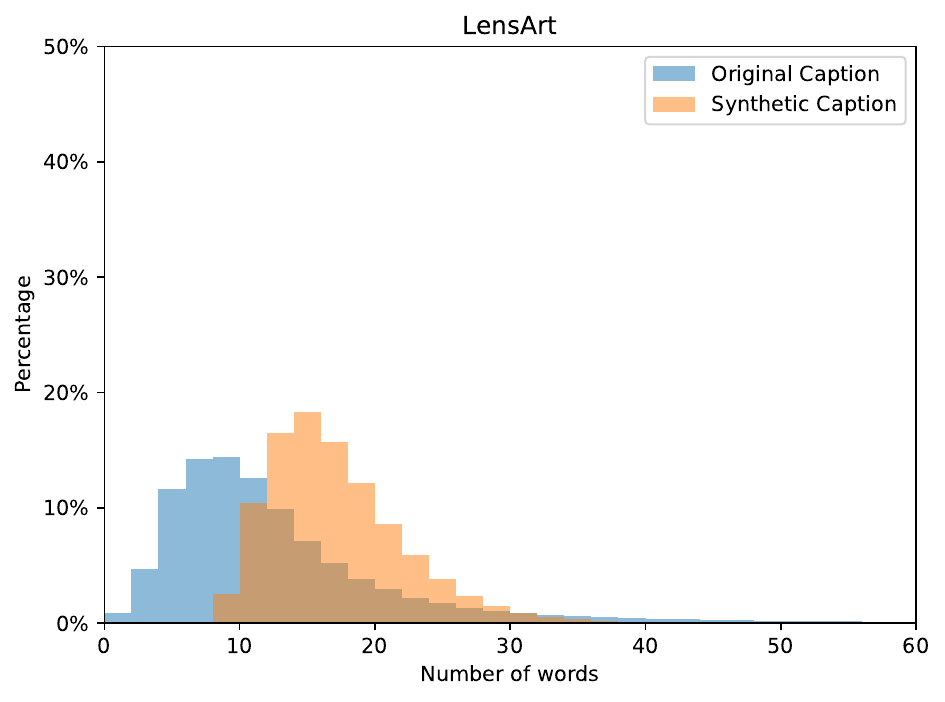}
\includegraphics[width=0.4\linewidth]{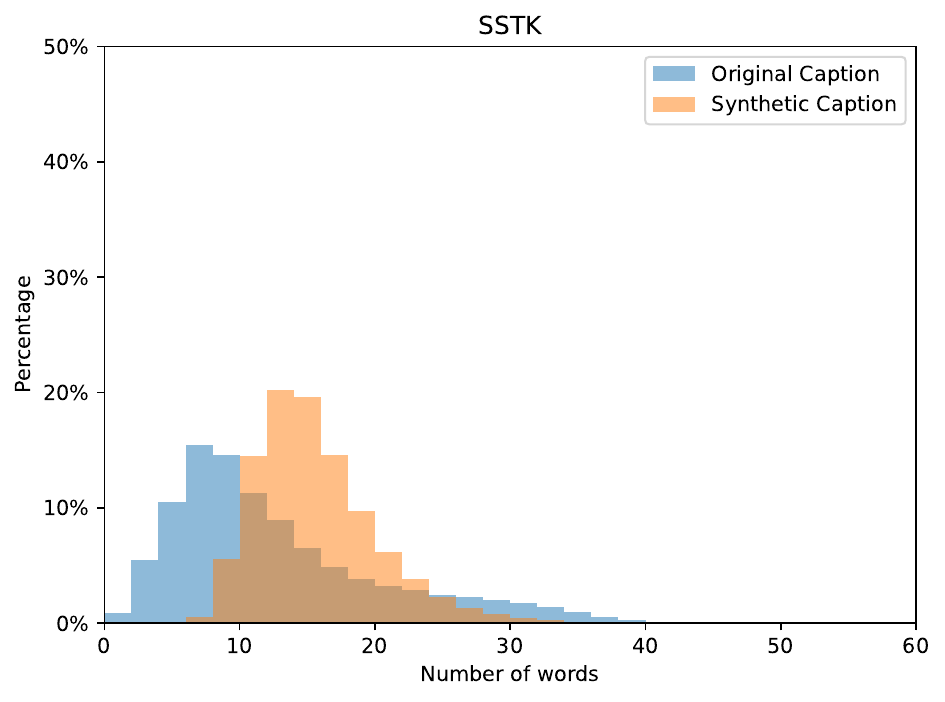}
\includegraphics[width=0.4\linewidth]{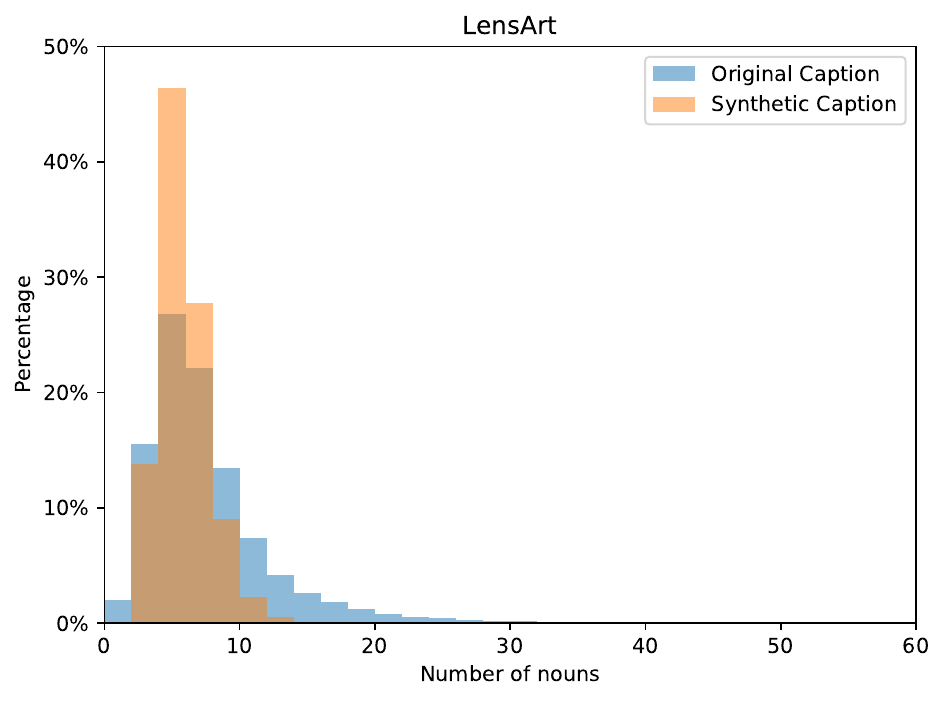}
\includegraphics[width=0.4\linewidth]{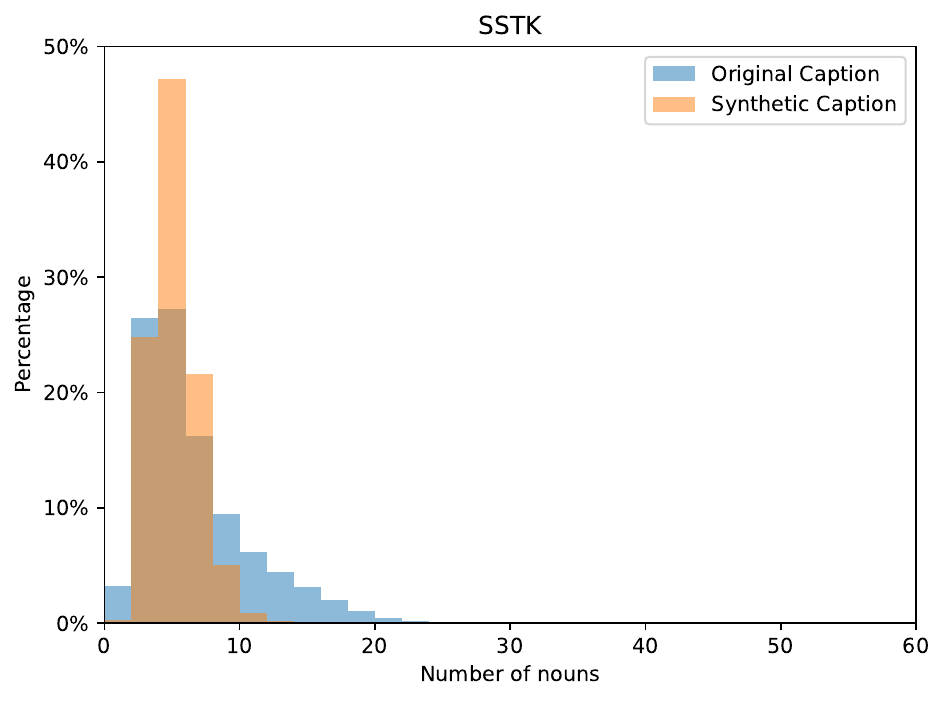}
\caption{Histograms of word and noun numbers in the original and synthetic captions of different datasets}
\label{fig:caption_hist}
\end{figure*}

For both LensArt and SSTK dataset, we present the histograms of number of words and nouns of original and synthetic captions respectively in Fig. \ref{fig:caption_hist}. Note that we overload the noun with noun and proper noun combined for simplicity. 
First, as shown in the first two figures, we see that synthetic captions are longer than original captions in terms of words, indicating augmenting original captions with synthetic captions can increase the supervision per image. Second, from the last two figures, we note that the number of nouns of synthetic captions are less than those in real captions on average. This is mainly caused by synthetic captions have less coverage in proper nouns, indicting the synthetic captions alone are not sufficient to train a generalist text-to-image model.

